\documentclass[11pt, a4paper, logo]{techreport}
\usepackage[authoryear, sort&compress, round]{natbib}
\usepackage{xspace}
\usepackage{multirow}
\usepackage{diagbox}
\usepackage{longtable}
\usepackage{ragged2e}
\usepackage{adjustbox}
\usepackage{array}
\usepackage{subcaption}
\usepackage{setspace}
\usepackage{algorithm}
\usepackage{algpseudocode}
\usepackage{listings}
\usepackage[bottom]{footmisc}
\usepackage{placeins}
\usepackage{tikz}

\addto\extrasenglish{
}

\makeatletter\@ifundefined{iflatexml}{\newif\iflatexml\@ifundefined{lx@newline}{\latexmlfalse}{\latexmltrue}}{}\makeatother
\iflatexml\else
\orgname{InfiX AI}
\orgaddress{}
\logofile{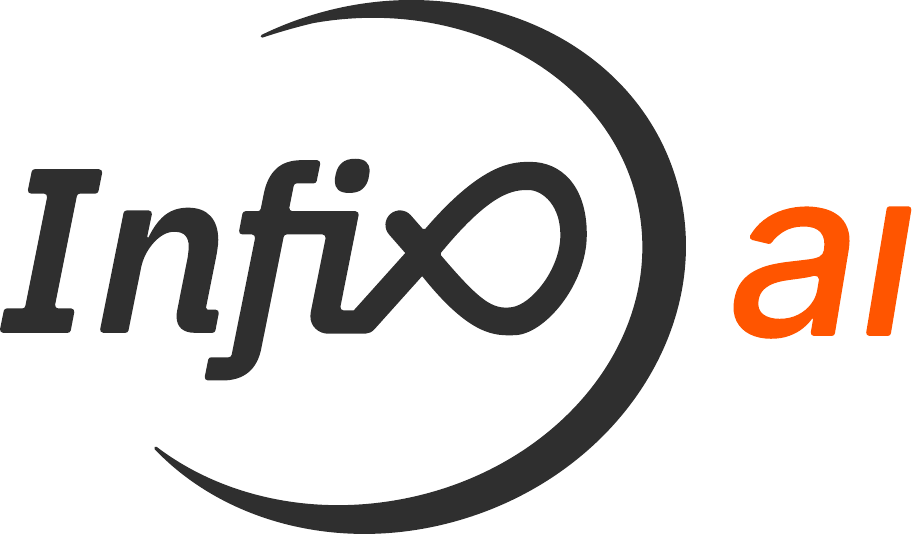}
\runningtitle{\modelname{}: New Progressive Training Paradigm towards Cost-Efficient Frontier Intelligence}
\reportnumber{001}
\paperurl{}                 % e.g. \paperurl{https://github.com/your-org/your-model}
\correspondingauthor{}      % e.g. \correspondingauthor{Corresponding author: name@example.com}
\fi
\newcommand{\modelname}{InfiX-V1}         % the model in general
\newcommand{\modelbase}{InfiX-V1-base}            % the pre-trained base model
\newcommand{\modelinstruct}{InfiX-V1-preview}     % the post-trained instruct model
\newcommand{\totalparams}{209B}
\newcommand{\activeparams}{14B}

\newif\ifnotes
\notestrue

\definecolor{medhi}{HTML}{FFE58A}\definecolor{medmid}{HTML}{F7C531}\definecolor{medlo}{HTML}{E09A12}
\definecolor{medrim}{HTML}{D68A0C}\definecolor{medstar}{HTML}{FFF3C4}
\definecolor{ribR}{HTML}{E0453A}\definecolor{ribB}{HTML}{3D6FD8}
\newcommand{\goldmedal}{\raisebox{-0.2em}{\tikz[baseline=0,x=0.86em,y=0.86em]{%
  \foreach \s in {-1,1}{%
    \begin{scope}[xscale=\s]
      \fill[ribB] (-0.42,1.18) -- (-0.30,1.18) -- (0.02,0.55) -- (-0.10,0.55) -- cycle;
      \fill[white] (-0.30,1.18) -- (-0.18,1.18) -- (0.14,0.55) -- (0.02,0.55) -- cycle;
      \fill[ribR] (-0.18,1.18) -- (-0.06,1.18) -- (0.26,0.55) -- (0.14,0.55) -- cycle;
    \end{scope}}
  \fill[medrim] (0,0.36) circle (0.37);
  \shade[top color=medhi, bottom color=medlo, middle color=medmid] (0,0.36) circle (0.32);
  \begin{scope}[shift={(0,0.36)}]
    \fill[medstar] (90:0.21) -- (126:0.085) -- (162:0.21) -- (198:0.085) -- (234:0.21) -- (270:0.085) -- (306:0.21) -- (342:0.085) -- (18:0.21) -- (54:0.085) -- cycle;
  \end{scope}}}}

\title{\centering \modelname{}: New Progressive Training Paradigm \\ towards Cost-Efficient Frontier Intelligence}

\author[*]{
\vspace{-0.8cm}
InfiX AI\\\vspace{-0.6em}
{\normalfont\small\texttt{research@infix.io}}
\vspace{-0.9cm}
}

\begin{abstract}
We introduce \emph{PACE}~(Progressive Adaptive Cognitive Expansion), a new cost-effective training paradigm that adapts the learning pace to a model's evolving capacity rather than training with a fixed model architecture and data schedule from the start.
PACE mitigates late-stage learning slowdown from fixed model capacity through three co-designed pillars.
First, \emph{InfiGrow} incrementally expands model width and depth during pre-training; under matched compute budgets, our Growth Scaling Law experiments show that it consistently outperforms training with a fixed model scale.
Second, \emph{InfiCoScale} co-scales training data diversity, difficulty, and quality in step with each growth stage, avoiding wasted compute on data mismatched to current model capacity.
Third, \emph{InfiEPFlow} provides system-level expert-parallel load balancing scheme that jointly optimizes computation and communication, improving training efficiency by 8\% over force balancing while maintaining strict compute balance.
Through four successive growth steps under PACE, we train \modelname{}, a Mixture-of-Experts~(MoE) model with \totalparams{} total and \activeparams{} activated parameters at remarkably low cost.
\modelbase{} achieves state-of-the-art performance, surpassing DeepSeek-V4-Flash-Base and MiMo-V2.5-Base; following post-training, \modelinstruct{} outperforms DeepSeek-V4-Flash~(0731), and matches DeepSeek-V4.1-Flash~(1017) across several domains.
Ultimately, \modelname{} delivers roughly $8\times$ higher \emph{intelligence per compute} than the best competing model, demonstrating that frontier capabilities can be achieved far more efficiently on a paced training path.
\end{abstract}

\begin{document}

\maketitle

\newsavebox{\teaserleftbox}\newlength{\teaserleftht}\newlength{\teasertop}\newlength{\teaserbot}\newlength{\teaseraxes}
% teaser-left rows: rotated side label + figure
\newcommand{\teaserrow}[2]{\parbox[c]{6pt}{\centering\rotatebox{90}{\fontsize{5.5}{6}\fontfamily{lmss}\selectfont\bfseries\color{black!75}#1}}\hfill\parbox[c]{\dimexpr\linewidth-8pt}{#2}}
% HTML (arXiv/LaTeXML) cannot reproduce the measured-box layout below, so it uses a pre-assembled image of the same figure.
\iflatexml
\begin{figure}[h]
\centering
\includegraphics[width=\textwidth]{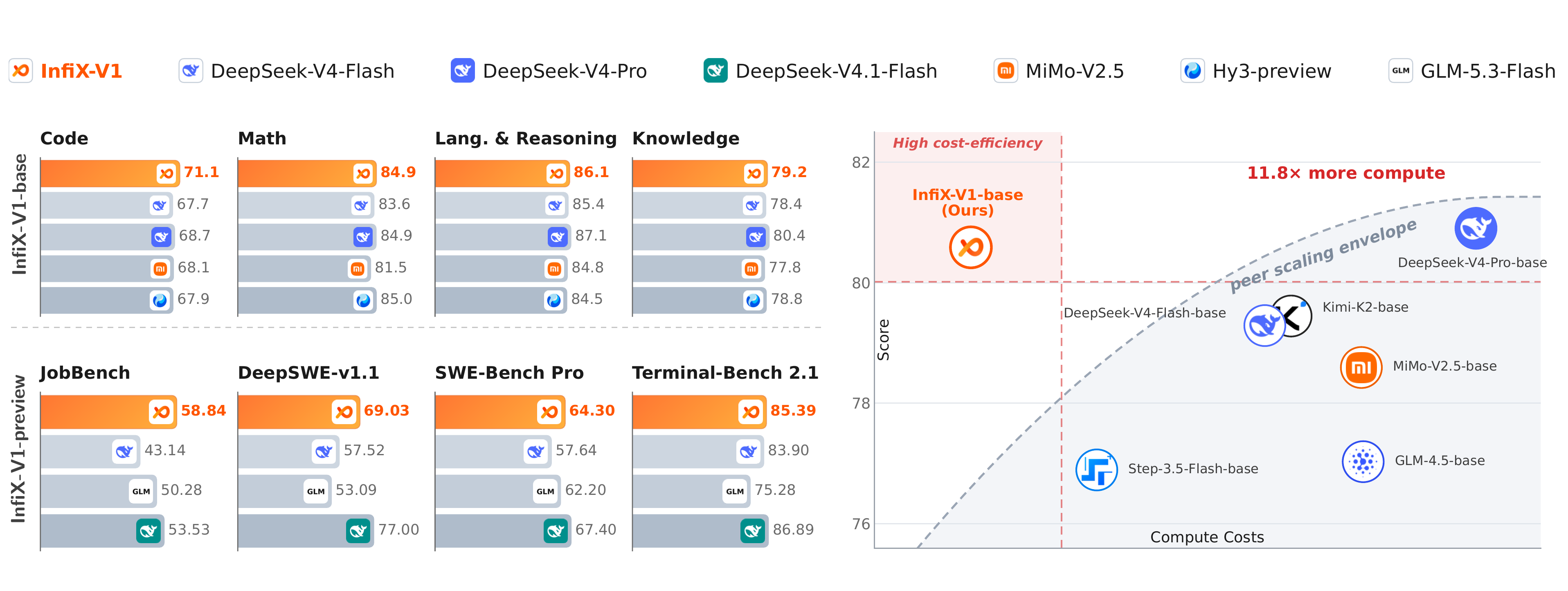}
\caption{
    \textbf{Top left}: benchmark performance of \modelbase{}.
    \textbf{Bottom left}: benchmark performance of \modelinstruct{}.
    \textbf{Right}: benchmark score versus pre-training compute.
}
\label{fig:teaser}
\end{figure}
\else
\begin{center}
\begin{minipage}{\textwidth}
\centering
% \vspace{-0.25cm}
% single-row legend on top (built from teaser-legend.pdf); old: \includegraphics[width=0.8\textwidth]{figures/teaser-legend.pdf} below the panels
\includegraphics[width=\textwidth]{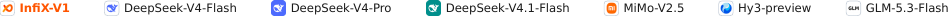}\\[0.6em]
\sbox{\teaserleftbox}{\begin{minipage}[c]{0.5035\textwidth}
% side labels + light dashed separator (instead of the "Base Model" / "Post-trained Model" header rules)
% old: \includegraphics[width=\linewidth]{figures/teaser-performance.pdf}\\[3pt]
% old: \includegraphics[width=\linewidth]{figures/teaser-posttraining.pdf}
\teaserrow{\modelbase{}}{\includegraphics[width=\linewidth]{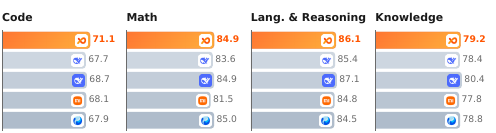}}\\[1pt]
{\color{gray!45}\xleaders\hbox to 3pt{\hss\rule[0.5ex]{1.6pt}{0.4pt}\hss}\hfill\null}\\[1pt]
% coauthor version: \includegraphics[width=\linewidth,trim=0 0 0 9.5pt,clip]{figures/teaser-posttraining.pdf}
\teaserrow{\modelinstruct{}}{\includegraphics[width=\linewidth]{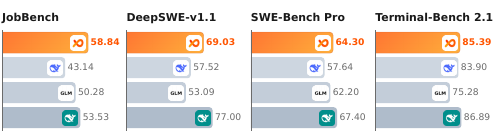}}
% % TODO: lower half of the teaser-left column reserved for a second figure
% \fcolorbox{gray!60}{gray!8}{\parbox[c][62pt][c]{\dimexpr\linewidth-2\fboxsep-2\fboxrule}{\centering\small\color{gray}placeholder}}
\end{minipage}}%
\setlength{\teaserleftht}{\ht\teaserleftbox}\addtolength{\teaserleftht}{\dp\teaserleftbox}%
% left panels (238pt-wide PDFs): cap line of the Code/Math titles is 6.69pt below the top, last grey bar ends 1.77pt above the bottom
\setlength{\teasertop}{\dimexpr(0.5035\textwidth-8pt)*669/23800\relax}%
\setlength{\teaserbot}{\dimexpr(0.5035\textwidth-8pt)*177/23800\relax}%
% right plot axes (1.1pt to 223.65pt of the 224.6pt-tall PDF) span the left panels from bar top to bottom line
\setlength{\teaseraxes}{\dimexpr\teaserleftht-\teasertop-\teaserbot\relax}%
% bottom-align: left panel bottom = x-axis of the right plot (x-axis at y=233.6 of the 234.6pt-tall teaser-efficiency-compact.pdf; axis labels moved onto the axes)
\resizebox{\textwidth}{!}{%
\raisebox{\dp\teaserleftbox}{\usebox{\teaserleftbox}}%
\hspace{0.015\textwidth}%
% teaser-efficiency-compact.pdf: built from teaser-efficiency.pdf without the footnote; Score vertical inside the middle of the y-axis, Compute Costs centred just above the x-axis
\raisebox{\dimexpr\teaserbot-\teaseraxes*95/22255\relax}{\includegraphics[height=\dimexpr\teaseraxes*22460/22255\relax]{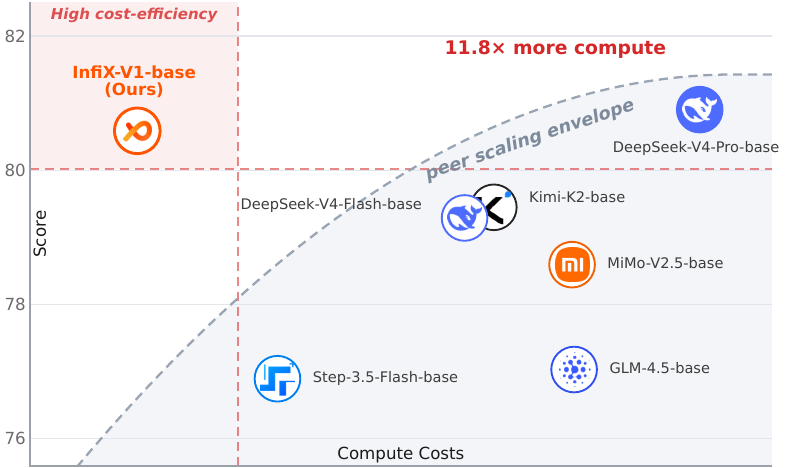}}%
}\\[0.3em]
\captionsetup{hypcap=false}%
\captionof{figure}{
    \textbf{Top left}: benchmark performance of \modelbase{}.
    \textbf{Bottom left}: benchmark performance of \modelinstruct{}.
    \textbf{Right}: benchmark score versus pre-training compute.
}
\label{fig:teaser}
\end{minipage}
\end{center}
\fi

\newpage

\begin{spacing}{0.9}
\setlength{\parskip}{0pt}
\tableofcontents
\end{spacing}

\newpage

\section{Introduction}
\label{sec:intro}

\begin{figure}[!t]
\centering
\includegraphics[width=\textwidth]{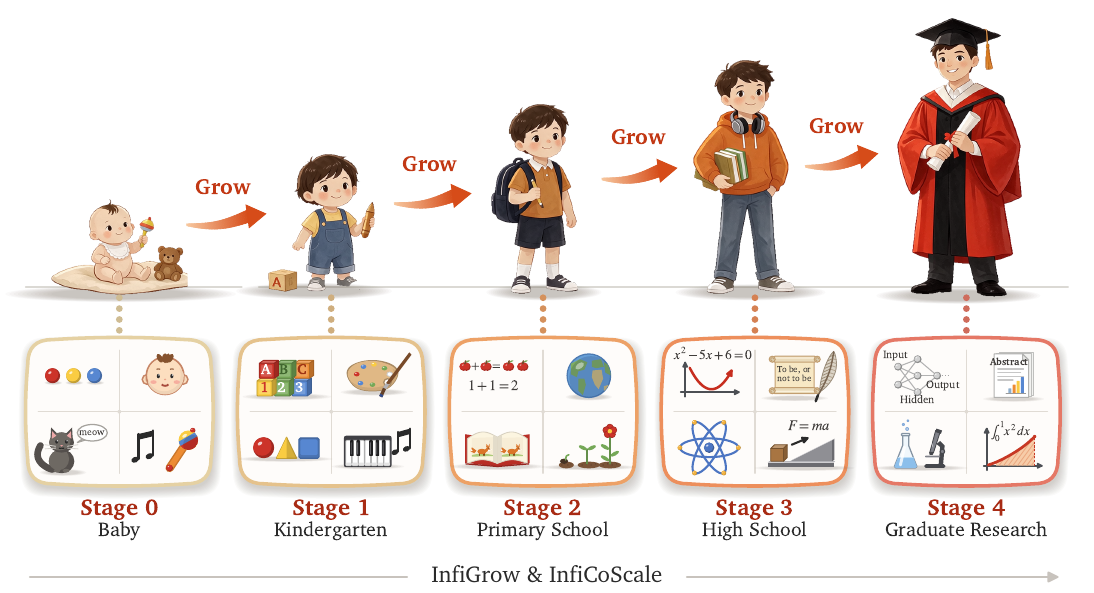}
\caption{\textbf{Learning like a growing mind.} As a person grows from a baby to a graduate researcher, both the capacity to learn and the difficulty of the knowledge rise stage by stage. \modelname{} follows the same principle: InfiGrow grows the model, 
and InfiCoScale moves the data towards more diverse and harder knowledge in step with each growth step.}
\label{fig:intelligence_growth}
\end{figure}

In recent years, large language models~(LLMs) have improved steadily as their parameters and training data grow.
As a result, the gap between open models and closed frontier models keeps narrowing~\citep{deepseekv3,deepseekv4}.
In particular, Mixture-of-Experts~(MoE) architectures have become the common choice for open models~\citep{deepseekv3,gptoss,mimov2flash}.
They enlarge the total number of parameters while keeping the computation per token small.
As models and datasets grow, the pre-training cost keeps rising.
In contrast, biological intelligence does not begin at full capacity; the human brain expands its functional capacity progressively as learning advances, matching cognitive development to increasingly complex skills.
At the same time, the knowledge a person learns becomes harder and more diverse.
This contrast motivates a similar growth principle for model training: \modelname{} introduces \emph{PACE}~(Progressive Adaptive Cognitive Expansion), a new training paradigm that flexibly adapts the learning pace to the model's evolving cognitive capacity.
By overcoming the late-stage learning slowdown under fixed model capacity, expanding model capacity in time, raising data complexity and diversity in lockstep, and eliminating redundant computation and communication overhead, PACE keeps every stage within a challenging yet feasible \emph{efficient learning zone}, maximizing intelligence gain per unit of compute.
Specifically, PACE is realized through three co-designed pillars:

The first pillar is \emph{InfiGrow}, a progressive model-growth mechanism in which the model expands step by step during pre-training~(\autoref{fig:intelligence_growth}).
Instead of training the target model from scratch, we first train a smaller model and then expand its width and depth in successive stages~\citep{gstack,solar,orthogonalgrowth,expertupcycling}.
For width expansion, we increase the number of activated experts per token and add new routed experts, which are copied from the most frequently selected experts with added noise.
For depth expansion, we stack layers while preserving the interleaving ratio of sliding window attention and full attention.
Growth experiments show that, under the same compute budget, InfiGrow outperforms fixed-scale training at every model size, and that width and depth growth are complementary when combined~(\autoref{sec:upscale}).

As the model scales in both layers and routed experts, training increasingly demands memory and communication efficiency; accordingly, our infrastructure optimizations target these two aspects.
During training, dynamic routing leads to imbalanced expert loads across devices.
To address this imbalance, existing systems typically force balanced computation and overlook the communication caused by token dispatching.
We therefore develop InfiEPFlow, an online and communication-aware EP load-balancing system.
InfiEPFlow maintains strict computation balance in every micro-batch and reduces the intra-node and inter-node communication of token dispatching under this constraint.
To achieve this, it repurposes the replica slots reserved for compute balance to shrink the communication domain of each token.
Compared with forced balanced training, InfiEPFlow improves training efficiency by over 8\%~(\autoref{sec:infra:epbalance}).

Just as a learner progresses to harder material as ability increases, the training data should co-evolve with the model.
A small model struggles to effectively learn from difficult data, while a large model that continues training on easy data wastes compute.
We therefore design InfiCoScale, which scales the training data alongside each growth stage.
Width expansion incorporates more long-tail and knowledge-rich data to support expert specialization, while depth expansion introduces more complex mathematics and code to strengthen multi-step reasoning.
In addition, later training stages prioritize data of higher quality.

Combining the three designs, we grow \modelname{} in four stages during pre-training into an MoE model with \totalparams{} total and \activeparams{} activated parameters.
\modelbase{} outperforms DeepSeek-V4-Flash-Base and MiMo-V2.5-Base, and after post-training, \modelinstruct{} outperforms DeepSeek-V4-Flash~(0731)~(\autoref{fig:teaser}).
We further define \emph{intelligence per compute}~(IPC) as a measure of how efficiently a training recipe converts compute into capability~(\autoref{sec:pretrain:ipc}).
By this metric, \modelbase{} achieves roughly 8x the IPC of the second best among all compared models.
These results show that a frontier-level model does not have to start at its final size: like human intelligence, it can grow across training stages and become more capable with much less compute.

\section{Growth Scaling Law}
\label{sec:upscale}

\begin{figure}[!t]
\centering
\includegraphics[width=\textwidth]{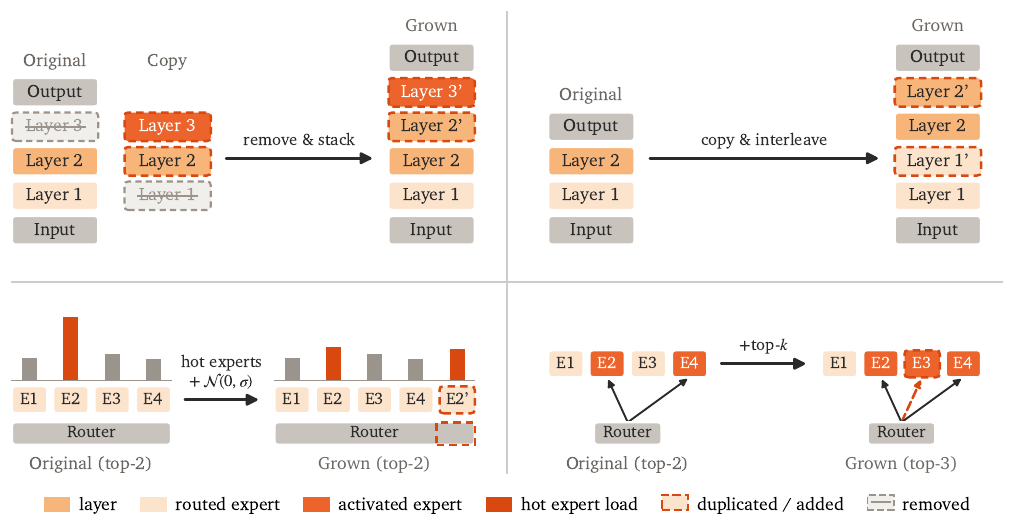}
\caption{Depth and width growth methods. \textbf{Top left}: stacking. \textbf{Top right}: interposition. \textbf{Bottom left}: hot-expert expansion. \textbf{Bottom right}: top-$k$ increase.}
\label{fig:upscale_methods}
\end{figure}

\paragraph{From scaling law to growth scaling law.}
Neural scaling laws show that, for a fixed model size, the loss approaches a floor set by model capacity as training proceeds, so additional tokens yield diminishing returns; meanwhile, the compute-optimal model size grows steadily with training compute~\citep{kaplan2020scaling,chinchilla}.
A fixed-size model is therefore compute-optimal at only one point during training: it is too large early on and too small later.
Ideally, the model should be grown as it learns, much like the way human intelligence develops.
From infancy to adulthood, the brain keeps growing and forming new neural connections, and the knowledge it takes in becomes harder as it grows, from colors and sounds for a baby to new research for a graduate student.
At every stage, the capacity of the learner and the difficulty of the knowledge rise together.
Under the PACE paradigm, \modelname{} grows the model and coscales its data together during pre-training through two coupled designs: \emph{InfiGrow} and \emph{InfiCoScale}.
InfiGrow trains the current model until its gains begin to slow, grows it, and then continues training the larger model, so that model size keeps pace with the growing compute budget.
Just as a growing person both knows more and reasons more deeply, the model can be grown in two directions.
Width growth adds experts and activated experts, which enlarges the capacity for domain knowledge, since more and finer-grained experts allow each expert to specialize in more focused knowledge~\citep{deepseekmoe}, and adding experts improves memorization more than reasoning~\citep{jelassi2024parrots}.
Depth growth adds layers, which strengthens reasoning, since deeper models perform better on multi-step and compositional reasoning at a similar parameter count~\citep{ye2024physics21,petty2024depth,saunshi2025looped}.
Together, width and depth growth expand the knowledge and the reasoning of the model step by step, so that InfiGrow follows the way human intelligence develops while saving significant compute.
% \modelname{} is grown four times along both directions~(\autoref{fig:upscale_process}).
% In step with each growth step, InfiCoScale adapts the training data to the new capacity: a width growth brings in more domain knowledge, and a depth growth brings in harder and higher-quality data, so that the larger model learns what it has just become ready to learn~(\autoref{sec:pretrain:orchestration}).
% Model growth has been explored in prior work~\citep{gstack,flm101b,orthogonalgrowth,sparseupcycling,dropupcycling,expertupcycling,longcatflash}, but these methods have mostly been studied at research scale and typically grow the model only once or along a single dimension.
% The only production-scale attempt, LongCat-Flash~\citep{longcatflash}, grows the model just once, in depth, to initialize pre-training. \modelname{} is the first large-scale MoE model pre-trained entirely with InfiGrow: it is grown four times, in both width and depth~(\autoref{fig:upscale_process}).
% Compared with training at a fixed size, this saves 20\% of the pre-training compute while achieving state-of-the-art performance.

%%======================================================================
\subsection{Depth Growth}
\label{sec:upscale:depth}

\paragraph{Method.}
Depth growth methods mainly fall into two categories: stacking and interposition~(\autoref{fig:upscale_methods}).
Stacking duplicates the whole model, removes the final $m$ layers of the original and the initial $m$ layers of the copy, and places the copy on top of the original, growing an $n$-layer model to $2(n-m)$ layers~\citep{gstack,solar}.
Interposition instead duplicates each layer and inserts the copy right after it, growing an $n$-layer model to $2n$ layers, and better preserves the layer-wise weight norm pattern of a converged model~\citep{orthogonalgrowth}.
However, interposition breaks the SWA/FA interleaving pattern of hybrid attention models, and the ablation study shows no clear gain from it over stacking.
Therefore, depth growth in the rest of this report refers to stacking, which is also the method adopted by \modelname{}~(\autoref{sec:pretrain:upscale}).

\paragraph{Experimental setup.}
\label{sec:upscale:setup}
To compare depth growth with training from scratch, scaling law experiments are performed on a series of MoE models~(\autoref{tab:upscale_models}).
All runs are trained on the same dataset with different numbers of training tokens.
Since \modelname{} is pre-trained with the warmup-stable-decay~(WSD) schedule~\citep{minicpm,hagele2024scaling}, the same schedule is used in these experiments.
Instead of training a separate run for every token budget, decay runs are branched from the stable phase at different token budgets, and the decay phase always takes 10\% of the total training steps, as recommended by~\citet{minicpm}.
The losses at the end of the decay runs, referred to as annealed losses, are used to fit the scaling laws.
Details of the model configurations and training are given in \autoref{app:upscale}.

\begin{figure}[!t]
\centering
\includegraphics[width=0.65\textwidth]{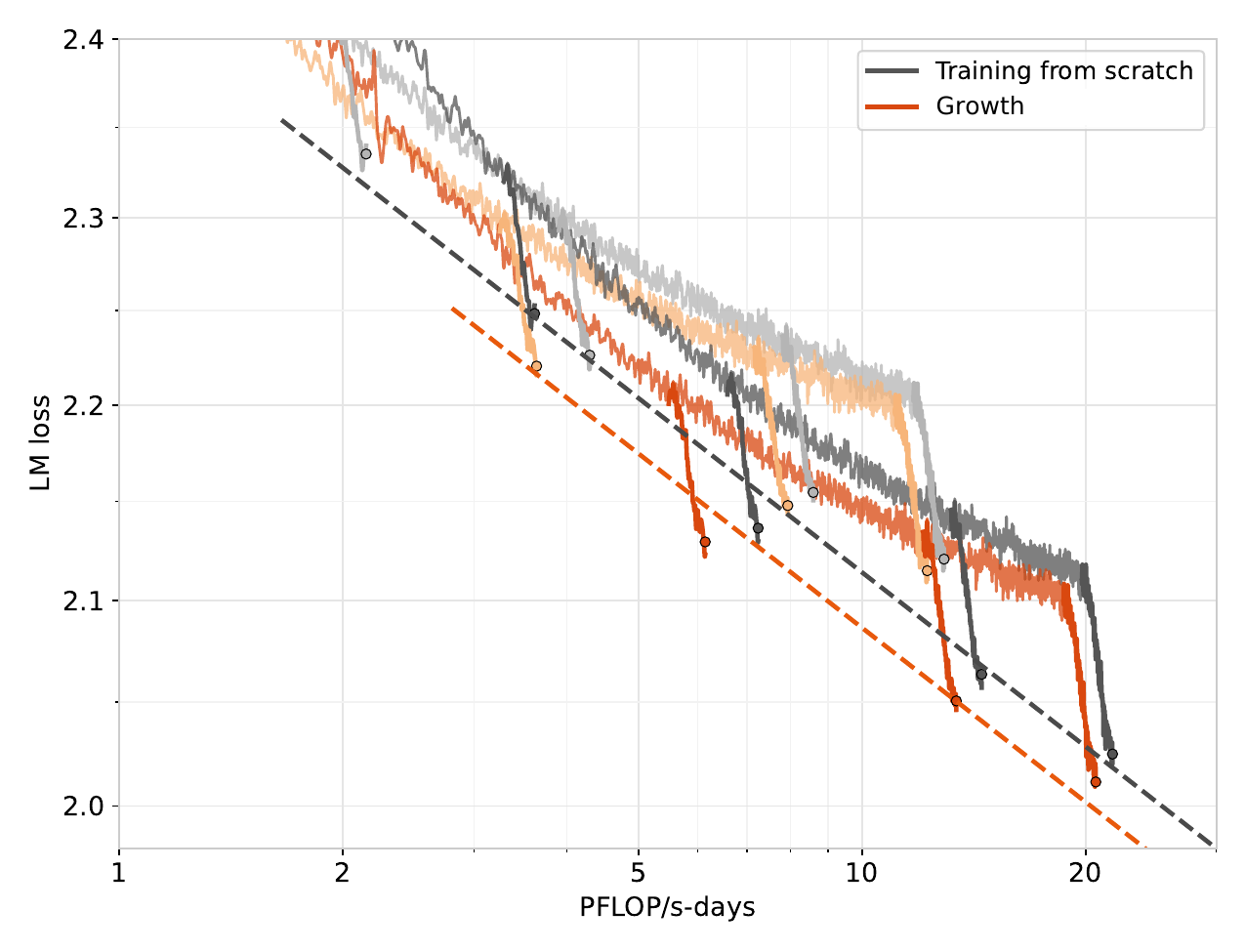}
\caption{Loss versus training compute for models trained from scratch~(grey) and depth-grown at a fixed step~(orange), with lighter and darker shades for the smaller and larger models, respectively. Thin lines show the training loss under a constant learning rate, thick lines show the decay runs, and the circles mark the annealed losses at the end of each decay. Dashed lines are power-law fits on the compute frontier with a shared exponent.}
\label{fig:upscale_law_depth}
\end{figure}

\paragraph{Depth growth law.}
\autoref{fig:upscale_law_depth} compares two ways of obtaining a model of the same final size.
In the baseline~(grey), the model is trained from scratch at its final size.
In depth growth~(orange), a shallower model is first trained and then grown to the same size as the baseline at the same point of training for all model sizes.
Both settings use the same data and learning rate schedule, and the training compute of the grown model includes the compute spent before the growth, so that the two settings are compared at equal compute.
The annealed losses on the compute frontier of both settings are fitted with a power law with a shared exponent $b$:
\begin{equation}
L = a\,C^{b},
\label{eq:upscale_law}
\end{equation}
where $L$ is the training loss and $C$ is the training compute in PFLOP/s-days. The fitted coefficients are listed in \autoref{tab:upscale_law_fit}.

\begin{table}[!t]
\centering
\caption{Fitted parameters of the depth growth law in \autoref{eq:upscale_law}, with $C$ in PFLOP/s-days.}
\label{tab:upscale_law_fit}
\begin{tabular}{l|c|c}
\toprule
 & Training from scratch & Depth growth \\
\midrule
LM loss & $2.426 \times C^{-0.0597}$ & $2.394 \times C^{-0.0597}$ \\
\bottomrule
\end{tabular}
\end{table}

A single depth growth shifts the whole frontier downward and reaches the same loss with about 20\% less training compute.

%%======================================================================
\subsection{Width Growth}
\label{sec:upscale:width}

\paragraph{Method.}
Width growth enlarges an MoE model in two ways: expert expansion and activated-expert expansion~(\autoref{fig:upscale_methods}).
Expert expansion adds routed experts by duplicating the hot experts, i.e., the experts that receive the most tokens.
Under auxiliary-loss-free load balancing~\citep{lossfree,deepseekv3}, the routing bias of an overloaded expert is lowered, so the hot experts can be identified as those with the lowest bias.
Each copy is initialized from its original expert with Gaussian noise of standard deviation 0.01 added to the weights~\citep{orthogonalgrowth}, so that the copy can diverge from the original during training.
The same noise is also added to the router weights of the copy, while no noise is added to its routing bias.
Activated-expert expansion instead raises the top-$k$ of the router, so that each token activates more of the existing experts without adding any parameters.

\begin{figure}[!t]
\centering
\includegraphics[width=\textwidth]{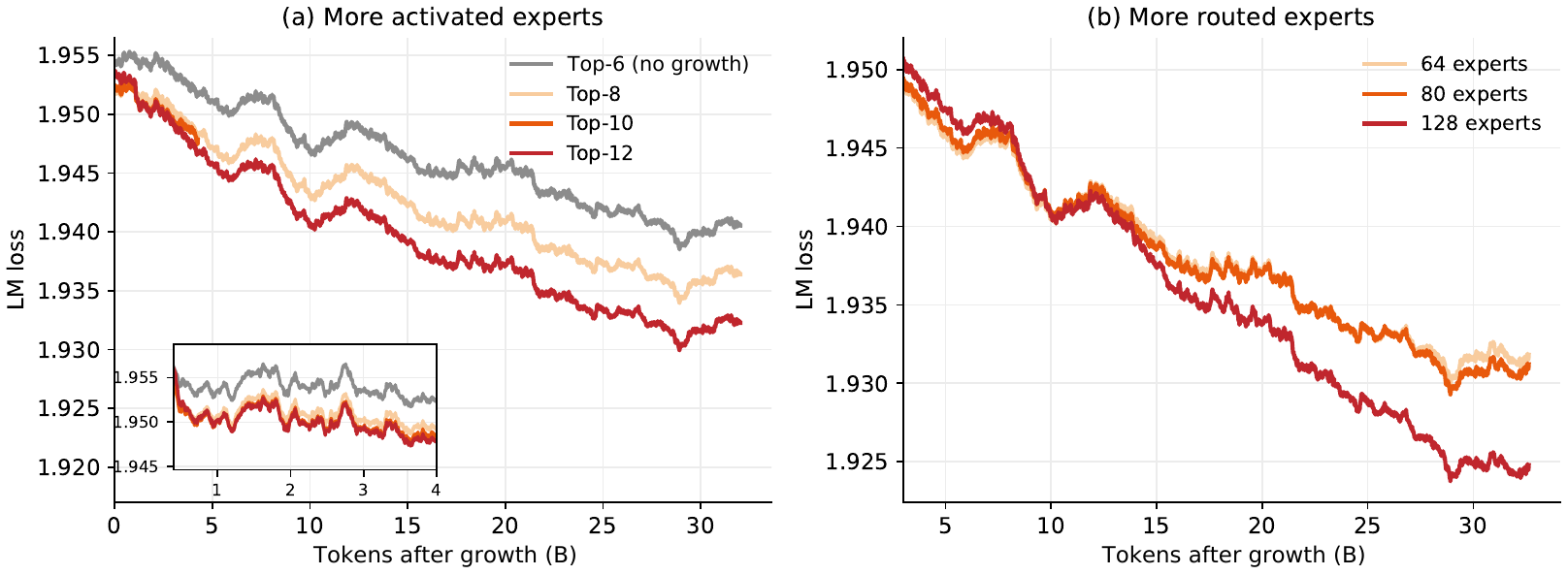}
\caption{Training loss after width growth of a 16B-A3B MoE model. \textbf{Left}: increasing top-$k$ from 6 to 8, 10 and 12; the inset zooms in on the first 4B tokens. \textbf{Right}: expanding the routed experts from 64 to 80 and 128 with the fixed top-$k$. Curves show the 501-step moving average.}
\label{fig:upscale_law_width}
\end{figure}

\paragraph{Width growth law.}
To understand the two ways of width growth, we conduct ablation studies on a 16B-A3B MoE model, increasing the top-$k$ from 6 to 8, 10 and 12, and, with fixed top-$k$, expanding the routed experts from 64 to 80 and 128~(\autoref{fig:upscale_law_width}).
Increasing top-$k$ reduces the loss immediately after the growth but shows clear diminishing marginal returns: the first two additional activated experts bring the largest gain, and each further increase brings a smaller one, in line with scaling laws~\citep{kaplan2020scaling,chinchilla,clark2022unified}.
Expanding routed experts shows the opposite trend: the more experts are expanded, the larger the gain, but the gain takes more training tokens to emerge.

These observations guide the width growth of \modelname{}.
Since the gain of increasing top-$k$ diminishes quickly, \modelname{} increases the number of activated experts only by a small amount at each width growth.
The gain of expanding routed experts emerges only after long training, \modelname{} therefore should leave enough training tokens after each expansion of routed experts for the expanded experts to converge.

\begin{figure}[!t]
\centering
\begin{minipage}{0.49\textwidth}
\centering
\includegraphics[width=\textwidth]{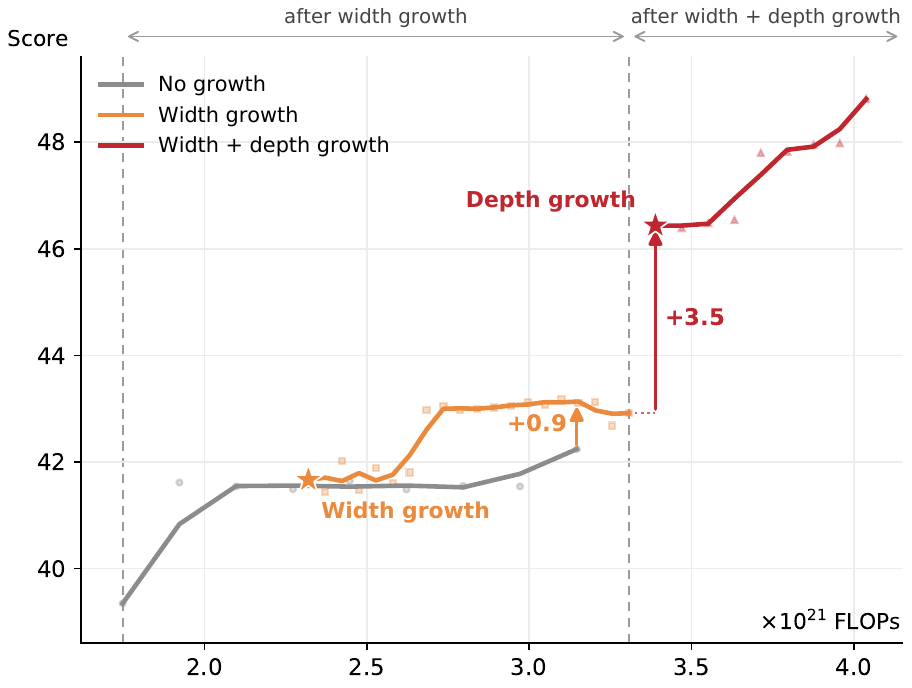}
\end{minipage}\hfill
\begin{minipage}{0.49\textwidth}
\centering
\includegraphics[width=\textwidth]{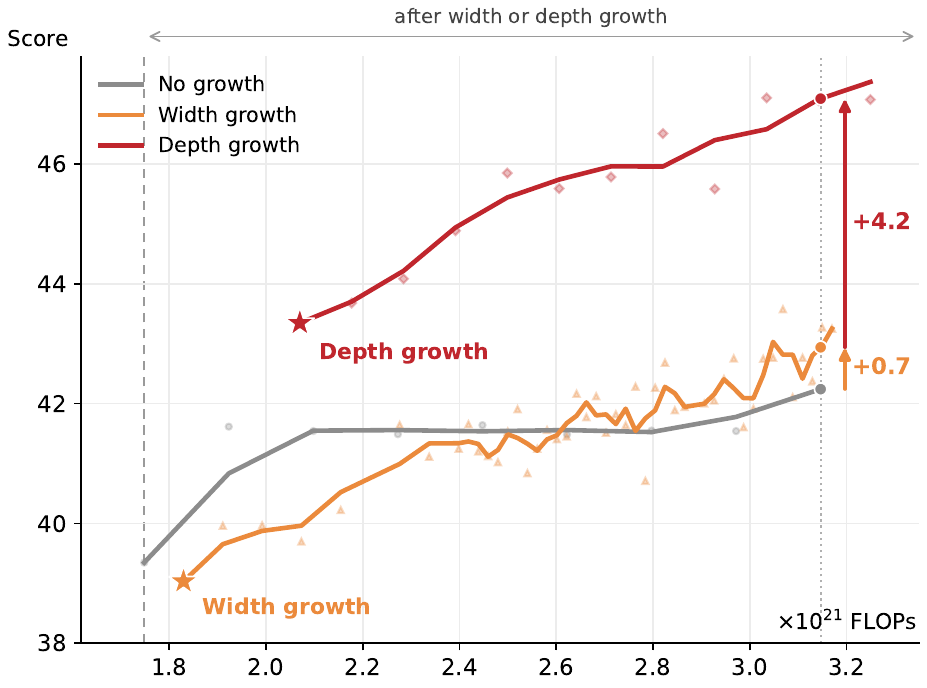}
\end{minipage}
\caption{Average score over the 11 benchmarks listed in \autoref{tab:ablation_upscale} in \autoref{app:ablation} versus training compute for the 16B-A3B MoE model. \textbf{Left}: width growth followed by depth growth. \textbf{Right}: width growth (top-$k$ 6$\to$8, routed experts 64$\to$72) and depth growth from the same checkpoint. Lines show the 3-checkpoint moving average, faint markers show the raw scores, and stars mark the growth points.}
\label{fig:upscale_bench}
\end{figure}

\paragraph{Width \& depth.}
\label{sec:upscale:bench}
\autoref{fig:upscale_bench} compares the average score of the model with and without growth at the same training compute.
As shown on the right, both width and depth growth outperform the model without growth, and depth growth brings a much larger gain than width growth: growing depth improves the average score by 4.8 points, while increasing top-$k$ from 6 to 8 and the number of routed experts from 64 to 72 improves it by 0.7 points.
Moreover, the two directions are complementary when applied progressively.
As shown on the left, a width growth first improves over the model without growth by 0.9 points at the same compute, and a subsequent depth growth brings a further gain of 3.5 points right after the growth.
% The detailed scores on each benchmark further show that growing width first and then depth outperforms depth growth alone, especially on math and code~(\autoref{tab:ablation_width_depth} in \autoref{app:ablation}), demonstrating the effectiveness of InfiGrow.

\FloatBarrier

\section{Infrastructure}
\label{sec:infra}

\begin{figure}[!t]
\centering
\includegraphics[width=\textwidth]{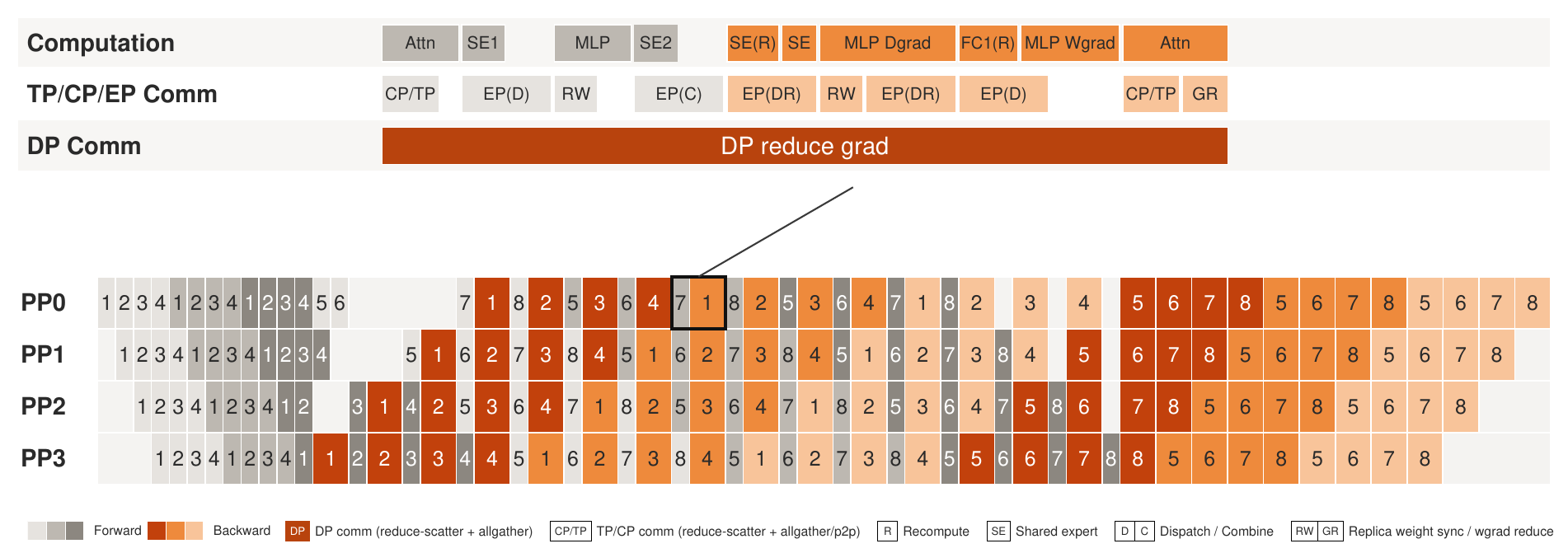}
\caption{Interleaved 1F1B pipeline schedule (PP${}=4$, VPP${}=3$, 8 micro-batches) with expert weight replication and overlapping computation and communication.}
\label{fig:pp_schedule}
\end{figure}

\modelname{} is trained with an in-house training framework built on Megatron-LM~\citep{megatron}.
The infrastructure is designed for the scenario of limited cluster scale, which raises two challenges:
(i)~\emph{memory}: the parameters, gradients, optimizer states and activations must fit into the memory of each GPU, and dynamic MoE routing can cause sudden memory spikes when many tokens are sent to a few experts; and
(ii)~\emph{communication}: with limited intra-node and inter-node bandwidth, the communication of CP, EP and DP cannot always be hidden behind computation.
An overview of the training pipeline schedule is given in \autoref{fig:pp_schedule}.

%%======================================================================
\subsection{General Infrastructure}
\label{sec:infra:general}

\subsubsection{Expert Parallelism Load Balance Strategy}
\label{sec:infra:epbalance}

System-level expert-parallel load balancing (EPLB) mitigates compute imbalance by redistributing expert computations across ranks. Online systems such as UltraEP~\citep{wei2026ultraep} and MoonEP~\citep{moonep2026} have made substantial progress in balancing computation.
Under dropless routing with homogeneous experts, the total expert computation per microbatch is invariant to the routing distribution.
For an EP group with $E$ experts evenly distributed across $R$ ranks, MoonEP shows that allocating $E/R$ additional replica slots per rank is sufficient to balance expert computation exactly across all ranks.
This guarantee prevents compute-load spikes and allows activation memory to be provisioned in advance.

As algorithmic balancing takes effect during training, the gains from further compute balancing diminish, yet EP communication can remain a bottleneck.
On bandwidth-limited clusters in particular, shared-expert and other independent computation may not fully hide A2A latency.
Unlike computation, communication traffic varies with expert placement and token assignment.
With communication deduplication in DeepEP~\citep{deepep2025}, co-locating selected experts or placing them closer to token sources can reduce A2A traffic even below that of conventional force balancing.
Inspired by this opportunity, we develop InfiEPFlow, an online EPLB system that optimizes communication while guaranteeing per-microbatch compute balance.

\paragraph{Online Planner.}
Using the same replica budget, InfiEPFlow partitions an EP group into two equal-sized communication domains, each hosting a complete expert set.
When computation can remain within the source domain, the number of reachable remote ranks per token decreases from $R-1$ to $R/2-1$.
Combined with destination-level deduplication, this reduces token A2A traffic even under uniformly random routing.
For inter-node EP, InfiEPFlow favors domains aligned with RDMA rails to minimize cross-domain traffic.
Across different stages of \modelname{} scaling, with varying expert counts, top-$k$ values, and EP sizes, domain contraction is estimated to reduce bottleneck communication cost by 30--50\% relative to conventional force balancing with uniformly distributed token destinations.

This method is not always compute-feasible, for instance, when an expert's workload exceeds twice the per-rank compute capacity.
In practice, algorithmic balancing (\autoref{sec:pretrain:setup}) and global shuffling
(\autoref{sec:pretrain:orchestration:global-shuffling}) reduce load skew, creating favorable conditions for domain contraction.

InfiEPFlow employs a low-latency GPU-resident planner to optimize replica placement and token assignment.
It first constructs an initial plan through two-level placement: the outer search partitions ranks into communication domains, while an LPT-based inner algorithm places replicas and assigns computation quotas.
For inter-node EP, we prune the outer search space based on network topology and evaluate the remaining candidates in parallel across SMs.
The planner then repairs compute-load violations and refines token assignments using the communication cost model.
When necessary, it adjusts replica placement, relaxes domain boundaries, or falls back to a compute-balanced placement under the same replica budget.

\paragraph{EP Communication Cost Model.}
InfiEPFlow models communication cost as the maximum transfer time across endpoints:
\begin{equation}
    \hat{T}_{\mathrm{comm}}(P,A)
    =
    \max_{h\in\mathcal{H}}
    \frac{V_h(P,A)}{B_h},
\end{equation}
where $P$ and $A$ denote replica placement and token assignment, $\mathcal{H}$ is the set of NVLink and RDMA send/receive endpoints, $V_h$ is the deduplicated communication volume at endpoint $h$, and $B_h$ is its measured effective bandwidth.
Concentrating replicas on the same rank or node can shift the communication bottleneck from network transfers to local reductions, as SM resources for communication are limited.
The resulting traffic patterns may also introduce channel imbalance and pipeline stalls, particularly in inter-node communication.
We optimize the inter-node communication path of DeepEP ~\citep{deepep2025} to address these bottlenecks and improve the accuracy of cost model.

\paragraph{Low-Overhead Replica Execution.}
As illustrated in \autoref{fig:pp_schedule}, InfiEPFlow decouples replica migration from EP token communication and schedules it independently of token dispatch and combine.
It prioritizes master expert execution while pipelining per-expert replica migration with expert computation, so each replica can start when its weights are ready.
Replica gradient reductions are chunked and overlapped with computation windows to minimize contention with attention TP/CP communication. 
InfiEPFlow is effective at large per-rank token counts, where A2A savings outweigh the largely fixed replica migration overhead.

\subsubsection{Memory- and Communication-Efficient Training}
\label{sec:infra:memcomm}

\paragraph{Memory-Efficient Muon.}
\label{sec:infra:muon}

Unlike element-wise optimizers such as AdamW, Muon~\citep{muon,moonlight} orthogonalizes the update of each weight matrix with Newton--Schulz iterations, which requires the complete matrix instead of a flat shard of it.
Distributed Muon therefore assigns each complete parameter to a single data-parallel~(DP) rank, which performs its orthogonalization, and the updated parameters are then gathered to all ranks~\citep{moonlight,megatronlayerwise}.
Since parameters differ in size, the shards owned by different ranks are uneven.
Standard collective communication, such as all-gather and reduce-scatter, requires a buffer of the same size on every rank, so each shard is padded to the largest one, and the communication buffer grows to the DP size times the largest shard, which significantly increases memory usage.
Avoiding the padding with uneven communication in NCCL, on the other hand, requires many SMs to reach a high bandwidth, which competes with computation.
\modelname{} designs uneven P2P-based communication operations that transfer only the actual shard of each rank, which removes the additional memory cost and reduces the communication cost.

\paragraph{Fine-Grained Recomputation across Pipeline Ranks.}
Under interleaved 1F1B pipeline parallelism~(PP)~\citep{narayanan2021efficient}, earlier PP ranks hold the activations of more in-flight micro-batches~(\autoref{fig:pp_schedule}).
Applying the same recomputation strategy to all PP ranks therefore either causes OOM errors on the first ranks or wastes memory and computation on the last ones.
\modelname{} applies recomputation at the granularity of individual modules within each layer, and assigns a different recomputation strategy to each PP rank.
This avoids OOM errors and balances the memory usage across PP ranks, while keeping the recomputation overhead low.

\paragraph{Communication-Efficient Context Parallelism in Sliding Window Attention.}
\label{sec:infra:cp}

Long-context training of \modelname{} relies on context parallelism~(CP).
However, widely used CP implementations~\citep{ringattention,transformerengine} are designed for FA and introduce redundant communication for SWA: ring attention and all-gather CP still pass the KV of the whole sequence across ranks, although each SWA query only needs the last $W$ tokens.
For SWA layers, \modelname{} replaces ring attention with a one-hop neighbour exchange.
With the load-balanced chunk layout of ring attention, the last $W$ tokens before each local chunk are always held by the two neighbouring ranks, so each rank only sends the last $W$ tokens of KV of its local chunks to its neighbours in a single round, computes the attention of each chunk locally in one kernel call, and sends back only the gradients of these tokens in the backward pass.
As a result, the communication of SWA layers no longer grows with the sequence length, and the sequential ring steps and the extra KV buffers are removed.
Since 30 of the 42 attention layers in \modelname{} are SWA layers, most of the CP communication is eliminated.
The attention sink is also supported by passing it directly to the attention kernel.

%%======================================================================
% refer to post-training infra
\subsection{Post-Training Infrastructure}
\label{sec:posttrain:infra}

\subsubsection{Sandbox for Agentic RL}
\label{sec:posttrain:sandbox}
To support large-scale agentic training, we built a scalable sandbox infrastructure in collaboration with our partners 
based on the Harbor framework and the E2B sandbox runtime. In this architecture, 
Harbor orchestrates multi-task environments and parallel rollouts at the upper layer, 
while E2B provides isolated execution sandboxes underneath. Our engineering optimizations focus on four core dimensions:

\paragraph{Strong Security Isolation.}
To mitigate risks from untrusted code execution, complex tool invocations, and environment state mutations during agentic training, each task runs inside a dedicated microVM-level isolated sandbox powered by E2B, enforcing strict boundaries across compute, file systems, and network stacks.
Combined with fine-grained network policies and credential isolation, this prevents single-point anomalies from affecting host machines or concurrent rollout instances, ensuring safe and controllable large-scale automated training.

\paragraph{Elastic Scaling and High-Concurrency Scheduling.}
Tailored to the bursty, fluctuating parallel sampling workloads in agentic reinforcement learning, we extended distributed dynamic auto-scaling and lifecycle management across the Harbor orchestration layer and the E2B sandbox pool.
The system automatically provisions, pools, and reclaims sandbox instances in batches according to real-time rollout concurrency demands, sustaining high-throughput trajectory collection while minimizing idle compute overhead.

\paragraph{High-Speed Image Pulling, Fast Cold Start, and Suspension.}
To address the frequent initialization overhead introduced by diverse task environments, we optimized the distribution pipeline for sandbox templates and environment images with multi-tier caching and on-demand pre-warming.
Through image layer reuse and node-level local caching, the system significantly reduces bandwidth contention and startup latency under high-concurrency scheduling, enabling rapid sandbox readiness.
Unfinished rollouts can also be suspended at safe interaction boundaries, with their sandbox and agent states preserved and the task-budget clock paused, and resumed after the policy update, so straggler trajectories never block training.

\paragraph{Decoupled Rollout Orchestration.}
A Rollout Hub decouples sandbox execution from GPU-side policy inference and optimization across network domains. It coordinates task-level rollout groups, allowing online replenishment and independent control of sandbox concurrency, and buffers complete requests before forwarding them across network boundaries; in large-payload stress tests, this reduces the delivery-failure rate to zero.

Benefiting from the high throughput and stability of this infrastructure, over 100{,}000 sandboxes were created and scheduled throughout training and evaluation, demonstrating its engineering reliability at scale.

\subsubsection{MXFP4 Training}
\label{sec:infra:rl}
\label{sec:infra:mxfp4}

Rollout dominates the cost of reinforcement learning~(RL), while quantizing a model only at deployment may degrade its performance online.
\modelname{} therefore adopts MXFP4~\citep{mxformat} quantization-aware training~(QAT)~\citep{jacob2018quantization} starting from the SFT stage.
Following gpt-oss~\citep{gptoss}, only the routed experts are quantized, as they account for the bulk of the parameters and memory bandwidth during decoding, while the other modules and all activations remain in BF16.
In the forward pass, their weights are quantized to MXFP4 with a block size of 32 and a power-of-two scale shared within each block, and then dequantized back; in the backward pass, the straight-through estimator~(STE)~\citep{ste} treats the quantization as the identity, since rounding has zero gradient almost everywhere.
In RL, the rollout engine performs MXFP4 inference to maximize throughput, and since rollouts are generated with quantized weights, QAT is also applied in the trainer, so that the log-probabilities computed by the trainer match those of the rollout engine more closely, reducing the training--inference mismatch.

\subsubsection{Efficient Full-Vocabulary Multi-Teacher On-Policy Distillation}
\label{sec:infra:opd}

We use multi-teacher on-policy distillation (MOPD)~\citep{gu2024minillm,lu2025onpolicy,kimik3,deepseekv4} to consolidate the capabilities of specialist models into a single student. As specialist training expands across task domains and reasoning budgets, keeping all teachers resident on GPUs becomes increasingly costly. Our MOPD pipeline consists of three stages: student rollout, teacher scoring, and student training. To operate within a limited GPU budget, these stages run sequentially on a shared GPU pool. At stage boundaries, inactive engines release GPU memory or offload model states to make room for the next stage. After each update, student weights are synchronized to the rollout engines for the next iteration.

During teacher scoring, work is distributed across a fixed number of data-parallel worker groups, each serving multiple teachers. To reduce switching overhead, we convert frozen teacher checkpoints offline into rank-local weight packs that match the inference engine’s parameter layout and sharding. Teachers with compatible configurations reuse an initialized engine by replacing its parameters, avoiding repeated initialization and conversion. Weight packs are loaded from distributed storage through reusable pinned-memory buffers. Upcoming teacher weights are prefetched while the current teacher is running, with a bounded prefetch capacity to control host-memory usage. A teacher-aware scheduler coordinates scoring across the worker groups. It balances the total number of tokens assigned to each group and schedules requests for the same teacher together to reduce repeated weight loading.

Full-vocabulary distillation requires retaining teacher supervision between scoring and student training. To reduce storage and communication costs, we cache the teachers’ final hidden states and reconstruct logits using the corresponding output heads during training~\citep{deepseekv4}. Mooncake Store~\citep{qin2025mooncake} provides shared storage for these hidden states, allowing supervision data to remain available after teacher engines release their GPU resources. The learner retrieves the data through RDMA and prefetches upcoming microbatches to overlap data transfer with computation.

Reconstructing logits on the learner requires memory for both teacher output heads and temporary activations. We shard teacher heads along the vocabulary dimension, cache the shards in bounded host memory, and load them onto GPUs according to the microbatch plan. Teacher-aware packing improves head reuse, and double-buffered transfers overlap head loading with computation. Head buffers remain valid until all dependent backward operations complete. To limit temporary logit storage, we compute output projections in token chunks and evaluate the full-vocabulary reverse-KL loss using specialized kernels. During backward, logits are recomputed to avoid retaining vocabulary-sized activations from the forward pass.

\begin{figure}[!t]
\centering
\includegraphics[width=\textwidth]{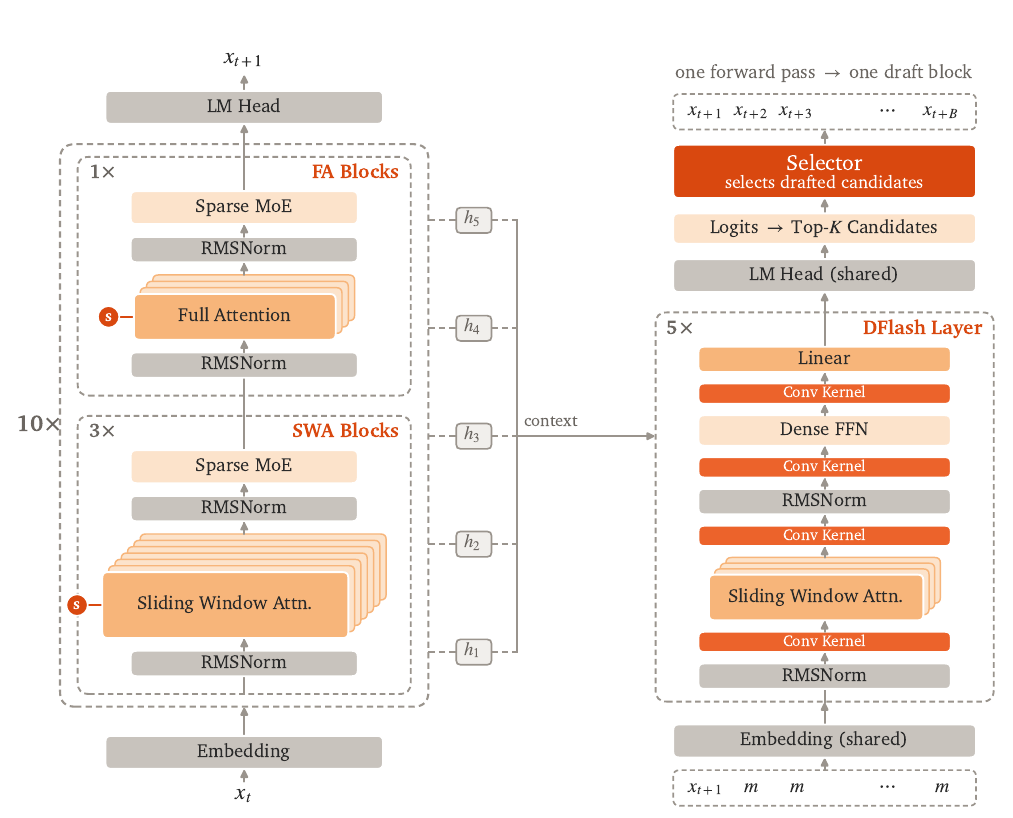}
\caption{Overall architecture of \modelname{}. \textbf{Left}: the backbone. In each attention layer, more boxes indicate more heads, a larger box indicates a larger QK head dimension, and $s$ denotes the attention sink. \textbf{Right}: the DFlash2 drafter, which takes the hidden states of the backbone as context, shares its embedding and LM head, and produces a block of draft tokens in a single forward pass.}
\label{fig:arch}
\end{figure}

\section{Model Architecture}
\label{sec:arch}

%%======================================================================
\subsection{Overview}
\label{sec:arch:overview}

\modelname{} follows a standard Transformer~\citep{transformer} backbone augmented with sparse MoE~\citep{switch,deepseekmoe} and hybrid attention~\citep{longformer,mistral,gptoss}.
Rather than being trained at its final size from scratch, \modelname{} is built through InfiGrow. 
The initial model should not be too deep, as a deeper model becomes much harder to converge after a depth growth~\citep{glorot2010understanding,wang2022deepnet}. Therefore, the initial model is designed to be short and fat.
The model is then grown four times, in width and in depth, to its final configuration of 42 layers and 288 routed experts with 13 activated per token (\autoref{sec:pretrain:upscale}).

\modelname{} is mainly composed of 10 hybrid blocks, each consisting of 3 consecutive sliding window attention~(SWA) layers followed by one full attention~(FA) layer, with a sliding window size $W$ of 512 to balance performance and efficiency on long-context tasks.
Both SWA and FA use Grouped-Query Attention~(GQA)~\citep{gqa}, with Rotary Positional Embedding~(RoPE) partially applied to the first 64 dimensions of query and key, and every attention layer is additionally equipped with a learnable attention sink (\autoref{sec:arch:hybrid:swa}) to improve performance and stability.
Except for the first layer, which uses a dense Feed-Forward Network~(FFN), all layers use a sparse MoE FFN.
The detailed configuration of \modelname{} is summarized in \autoref{tab:model_config}.

\begin{table}[!t]
    \centering
    \small
    \caption{Model configuration of \modelname{}.}
    \label{tab:model_config}
    \begin{tabular}{@{}l l c@{}}
    \toprule
    \textbf{Block} & \textbf{Configuration} & \textbf{Value} \\
    \midrule
    \multirow{2}{*}{\textbf{Backbone}}
    & Layers (Total/SWA/FA) & 42/30/12 \\
    & Hidden Size & 4096 \\
    \midrule
    \multirow{5}{*}{\textbf{Attention}}
    & SWA Heads (Q/KV) & 80/8 \\
    & SWA Head Dimensions (QK/V) & 192/128 \\
    & Sliding Window Size & 512 \\
    & FA Heads (Q/KV) & 64/4 \\
    & FA Head Dimensions (QK/V) & 128/128 \\
    \midrule
    \multirow{4}{*}{\textbf{MoE}}
    & Layers (Dense/MoE) & 1/41 \\
    & Experts (Routed/Shared/Activated) & 288/1/13 \\
    & Expert FFN Dimension & 1408 \\
    & Dense FFN Dimension & 15,488 \\
    \midrule
    \multirow{6}{*}{\textbf{DFlash2}}
    & Layers (Total/SWA/FA) & 5/5/0 \\
    & Hidden Size & 4096 \\
    & SWA Heads (Q/KV) & 64/8 \\
    & Sliding Window Size & 1024 \\
    & Head Dimensions (QK/V) & 128/128 \\
    & Dense FFN Dimension & 15,488 \\
    \bottomrule
    \end{tabular}
\end{table}

%%======================================================================
\subsection{Hybrid Attention}
\label{sec:arch:hybrid}

The cost of full attention grows quadratically with the sequence length in computation and linearly in KV cache, which dominates both long-context training and inference.
Sliding window attention restricts each query to a fixed number of the most recent tokens, and interleaving SWA with FA layers has been widely adopted to reduce this cost, from GPT-3~\citep{gpt3,sparsetransformer} and Mistral~\citep{mistral} to gpt-oss~\citep{gptoss}.
Meanwhile, studies show that language models tend to allocate excess attention onto a few initial tokens, known as the attention sink~\citep{streamingllm,sinkemerge}.
This problem becomes more severe in SWA, as the initial tokens quickly fall out of the window, and the hybrid SWA models suffer obvious performance degradation~\citep{mimov2flash}. The learnable sink addresses this and mitigate training instability.
\modelname{} interleaves SWA and FA layers at a ratio of 3:1, together with a learnable attention sink in every layer.

\subsubsection{Sliding Window Attention and Attention Sink}
\label{sec:arch:hybrid:swa}

Theoretically, Stacking $l$ SWA layers extends the receptive field to $l(W-1)$ tokens~\citep{mistral,mixtral}, and since the first layer of \modelname{} is an FA layer and every 3 SWA layers are followed by another FA layer, the input of every SWA layer has already aggregated the entire prefix.
For a single attention head, let $\mathbf{q}_i, \mathbf{k}_i \in \mathbb{R}^{d}$ be the query and key of the $i$-th token.
\modelname{} applies the same desigin of gpt-oss\citep{gptoss}, each head is equipped with a learnable attention sink $s \in \mathbb{R}$ that is added only to the denominator of the softmax, and the output of token $i$ is
\begin{equation}
    a_{ij} = \frac{\mathbf{q}_i \mathbf{k}_j^\top}{\sqrt{d}}, \qquad
    p_{ij} = \frac{\exp(a_{ij})}{\exp(s) + \sum_{j' \in \mathcal{S}_i} \exp(a_{ij'})}, \qquad
    \mathbf{o}_i = \sum_{j \in \mathcal{S}_i} p_{ij} \, \mathbf{v}_j,
    \label{eq:attn}
\end{equation}
where $\mathcal{S}_i$ is the set of visible tokens.
FA attends to the whole prefix, $\mathcal{S}_i^{\mathrm{FA}} = \{\, j \mid 1 \le j \le i \,\}$, while SWA attends only to the latest $W$ tokens, $\mathcal{S}_i^{\mathrm{SWA}} = \{\, j \mid i - W < j \le i \,\}$.
The sink is equivalent to a virtual token with logit $s$, and a head can put most of its weight on the sink token and output a vector close to zero when the hidden states is not intended to update.

In practice, however, naively replacing FA layers with SWA layers at a 3:1 ratio, even with an attention sink, still degrades performance (\autoref{tab:ablation_fa_swa} in \autoref{app:ablation}).
The experiments of \modelname{} show that this can be largely resolved with little extra cost by enlarging the capacity of SWA layers.
The ablation study on a 16B-A3B MoE model~(\autoref{tab:ablation_swa} in \autoref{app:ablation}) leads to:
(i)~more KV heads in SWA layers bring the hybrid model close to the full-attention baseline;
(ii)~a larger head dimension in SWA layers further surpasses the baseline, while a larger head dimension in FA layers brings little additional gain; and
(iii)~gated attention with more query heads brings no obvious gain over simply adding KV heads.
Based on these conclusions, \modelname{} increases the number of query heads, the number of KV heads and the head dimension only in SWA layers, while keeping FA layers unchanged~(\autoref{tab:model_config}).

%%======================================================================
\subsection{Draft Model}
\label{sec:arch:dflash}

To accelerate inference, \modelname{} is equipped with a lightweight drafter for speculative decoding, which follows the block diffusion design of DFlash~\citep{dflash} and its refinement DFlash2.
Instead of generating draft tokens one by one, the drafter takes the hidden states of the backbone and a clean anchor token as input, produces a block of 7 draft tokens in a single forward pass, and the backbone then verifies the whole block in parallel.
The drafter consists of 5 dense Transformer layers~(\autoref{tab:model_config}), all of which use SWA with GQA, with 64 query heads and 8 KV heads, and a learnable attention sink.
Tokens within a draft block attend to each other bidirectionally, while each block only sees the 1024 backbone positions right before its anchor.
On top of DFlash, DFlash2 introduces a lightweight selector that re-ranks the candidates at each block position into a top-16 shortlist, after a grouped dynamic convolution~(kernel size 2, group size 16) mixes the local context.

\section{Pre-Training}
\label{sec:pretrain}

\subsection{Data Construction}

% We collect a large-scale training corpus that is both high-quality and broad in domain coverage. Since our goal is efficient training, we apply rigorous filtering and cleaning to retain only the most informative data.
We collect a large-scale training corpus that is both high-quality and broad in domain coverage.
Following the development of human intelligence~(\autoref{fig:intelligence_growth}), we organize the training data by coscale stages, so that the data grows together with the model.
The base model first learns broad, basic world knowledge, which lays a solid foundation for later growth.
As its width and depth expand through InfiGrow, we gradually increase the diversity and difficulty of the data to make full use of the added capacity while keeping training efficient.

\subsubsection{Pre-Training Data Curation}
\paragraph{Web.} %我们基于抓取和开源数据收集了广泛的网页数据，主要来自FineWeb和Common Crawl。这些数据经过一个从粗到细的过滤pipeline。我们首先对这些网页数据根据领域和用途定义了分类taxonomy，对低质量语料类别进行粗筛除。我们又结合网页的类别定义筛除规则，进行规则过滤和页面元素去除。最后我们基于模型细粒度过滤低质量页面并根据领域和用途对数据平衡配比。在此基础上，我们对中文和英文网页维护了一个高质量url池，包括百科网站和问答网站等。这些高质量网页在通用解析器下往往无法完整保留正文内容。因此我们对每个站点设计了优化过的解析策略以提取这些正文、表格和latex格式等，避免提取到页面元素。对于高质量站点，我们除了进行通用基础质量检测，错误信息分类器也被引入用来过滤站点中的错误知识和内容，避免在预训练阶段引入过多错误信息。对于可修改的错误内容，我们利用强模型改写修复。
% We collect a broad range of web data from both our own crawls and open-source datasets, primarily FineWeb~\citep{Penedo2024FineWeb} and Common Crawl.\footnote{\url{https://commoncrawl.org/overview}} This data goes through a coarse-to-fine filtering pipeline. First, we define a taxonomy of web pages by domain and purpose, and discard categories dominated by low-quality content. Next, we apply category-specific rules to filter documents and remove page boilerplate. Finally, we use model-based classifiers to filter out low-quality pages at a fine-grained level, and rebalance the data across domains and purposes. Beyond this general pipeline, we maintain a pool of high-quality Chinese and English URLs, such as encyclopedia and question-answering websites. General-purpose parsers often fail to preserve the full main content of these pages, so we design an optimized parsing strategy for each website to extract the main text, tables, and LaTeX content while excluding page boilerplate. In addition to standard quality checks, we apply a misinformation classifier to these high-quality websites to filter out factually incorrect content, so that little erroneous knowledge is introduced during pretraining. Where the errors are correctable, we use a strong model to rewrite and fix the content.
We collect a broad range of web data from publicly available websites and open-source datasets.
This data goes through a coarse-to-fine filtering pipeline.
First, we define a taxonomy of web pages by domain and purpose, and discard categories dominated by low-quality content.
Next, we apply category-specific rules to filter documents and remove page boilerplate.
Finally, we use model-based classifiers to filter out low-quality pages at a fine-grained level, and rebalance the data across domains and purposes.
Beyond this general pipeline, we maintain a pool of high-quality Chinese and English websites.
General-purpose parsers often fail to preserve the full main content of these pages, so we design an optimized parsing strategy for each website to extract the main text, tables, and LaTeX content while excluding page boilerplate.
In addition to standard quality checks, we apply a misinformation classifier to these high-quality websites to filter out factually incorrect content, thereby limiting the introduction of erroneous knowledge during pretraining.
Where the errors are correctable, we use a strong model to rewrite and fix the content.

\paragraph{Math.} %高质量预训练数学数据非常稀缺\citep{liang2024trait,zhang2025automathtext,zhou2025megamath}。我们检查了现有流行的开源数据集，包括FineMath、MegaMath等，尽管这些数已经经过严格的质量过滤，但仍然存在大量低质量样本以及数学无关内容。为了解决这一问题，我们扩展了数学数据的收集范围包括Web Math，高质量开源数据集以及数学站点池。我们在通过了通用Web数据清洗的数据基础上应用fastText分类器识别数学文档，并构建优化过的解析器以保证公式和符号质量。数学站点池是基于种子站点池，通过我们搭建的的自动采集解析流水线不断扩展得到的。我们还通过向量检索从其他领域数据中召回数学相关语料。这些收集到的数据先经过严格的网页元素去除，再被送入规则和模型结合的过滤器中去除低质量样本，并进行去重。
% High-quality mathematical pretraining data is scarce at the scale needed to train large language models~\citep{liang2024trait,zhang2025automathtext,zhou2025megamath}. We examined popular open-source datasets such as FineMath~\citep{benallal2025smollm2} and MegaMath~\citep{zhou2025megamath}, and found that, despite rigorous quality filtering, they still contain a large number of low-quality samples and math-irrelevant content. To address this issue, we broaden our sources of mathematical data to include Common Crawl, high-quality open-source datasets, and a pool of math websites. Starting from data that has passed our general web cleaning pipeline, we apply a fastText classifier~\citep{joulin2017fasttext} to identify mathematical documents, and build an optimized parser to preserve the quality of formulas and symbols. The math website pool starts from a set of seed websites and is continuously expanded by our automated crawling and parsing pipeline. We additionally use vector retrieval to recall math-related content from data in other domains. All collected data first undergo thorough removal of web page boilerplate, then pass through a hybrid rule- and model-based filter that removes low-quality samples.
High-quality mathematical pretraining data is extremely scarce~\citep{zhang2025automathtext,zhou2025megamath,liang2024trait}.
Even rigorously filtered open-source mathematical corpora still contain a large number of low-quality samples and math-irrelevant content.
We therefore broaden our sources of mathematical data to include general web data, high-quality open-source datasets, and math-oriented websites.
Starting from data that has passed our general web cleaning pipeline, we apply a fastText classifier to identify mathematical documents, and build an optimized parser to preserve the quality of formulas and symbols.
We assemble the pool of math-oriented websites with our automated crawling and parsing pipeline, and additionally use vector retrieval to recall math-related content from data in other domains.
All collected data first undergo thorough removal of web page boilerplate, then pass through a hybrid rule- and model-based filter that removes low-quality samples.

\paragraph{Code.} %预训练代码数据主要分为Repo数据和Web Code数据。Repo数据来源是Github上的开源仓库和开源数据集，github数据的截断日期为2026年。这些Repo数据经过了严格的质量过滤，包括基于OpenCoder的规则过滤\citep{huang2025opencoder}和基于模型的质量过滤，去除了语法错误、非代码内容以及低质量样本。我们还通过规则去除了代码数据中的版权头以减少无效内容。我们还对代码数据根据应用场景定义了细粒度的类别标签，并基于标签和代码语言进行平衡。Web Code数据来源是Web站点池和基于Web数据和高质量开源数据集，我们优化了解析器并且训练了独立的质量过滤模型进行过滤。
Our code pretraining data consists of two parts: repository data and web code data.
The repository data is collected from publicly available code repositories, with a cutoff date of 2025-12, and from open-source code datasets.
We apply strict quality filtering to this data, combining rule-based filtering built upon OpenCoder~\citep{huang2025opencoder} with model-based quality filtering, to remove code with syntax errors, non-code content, and low-quality samples.
We also use rules to strip copyright headers, which carry little useful information.
In addition, we assign fine-grained category labels to the code based on its application scenario and balance the data across these labels and programming languages.
The web code data is collected from our website pool, general web data, and high-quality open-source datasets.
For this data, we use an optimized parser to extract the code and train a dedicated quality model to filter it.

\paragraph{Knowledge.} % 我们收集了书籍和论文以扩充科学和文学语料。论文包括PDF论文和latex源码，书籍由常见的电子书格式（如word,epub等）和PDF组成。对于PDF数据，我们搭建了一套高吞吐的OCR处理管线，将其转成纯文本数据。对于Latex源码，我们保留全部数学公式并展开解析正文内容。对于电子书文件，我们通过阅读器解析出正文，并设计策略转译公式为Latex格式。在上述的解析正文之后，所有书籍和论文语料都经过一个质量过滤模型以筛除低质语料。在去重时，为了控制长文本在minhash时资源消耗，我们设计了基于显著段落的计算签名方法，在低消耗的同时避免了清洗跨格式、版本重复书籍的资源浪费。
% We collect books and academic papers to enrich our scientific and literary corpus. The papers come as either PDF files or LaTeX sources, while the books come in common e-book formats, such as Word, EPUB, PDF, as well as many other formats. For PDF files, we build a high-throughput OCR pipeline that converts them into plain text. For LaTeX sources, we preserve all mathematical formulas and recursively resolve included files when parsing the main content. For e-book files, we extract the main text with a reader-based parser and design dedicated strategies to convert formulas into LaTeX. After content extraction, all book and paper data pass through a quality model that filters out low-quality samples. For deduplication, applying MinHash~\citep{broder1997resemblance} directly to such long documents is prohibitively expensive, so we compute signatures from the top-$k$ longest passages of each document instead. This keeps the cost low while still detecting duplicate books that appear in different formats or editions.

We collect publicly available books and academic papers across a broad range of subjects.
We parse these documents in their original formats to preserve content such as formulas and tables where present.
Reference lists are removed, since they consist of citation strings rather than content.
The parsed documents pass the same rule- and model-based quality filters as web data, and are labeled by domain and educational level so that STEM material can be upsampled in the mixture.

\subsubsection{Mid-Training and Long-Context Data Curation}

Mid-training sits between pre-training and post-training~\citep{liu2025midtraining,tu2025survey}.
While pre-training establishes broad representations on short contexts and post-training aligns the model through long supervised stages that can erode trainability~\citep{singhal2026priming,wang2025octothinker}, mid-training narrows the gap along three axes: distribution, shifting the data mixture toward the regions post-training will sample; format, exposing the model to agentic-style interaction with software artifacts to preserve trainability; and length, extending the context window over which attention is trained.

\paragraph{Complementary Recall Signals.}
The recalled portion of the mid-training mixture is drawn entirely from the pre-training corpus through three complementary signals.
Perplexity-based filtering keeps documents whose perplexity drops after continued pre-training of a reference model, rejecting increases and extreme drops; the retained band follows the pre-training distribution.
Tag-based filtering employs a trained scorer to label documents by subject, educational level, and difficulty, transforming the corpus into queryable buckets.
Embedding-based retrieval selects long-tail, knowledge-dense regions.
The three signals are combined with the rewritten and agentic-style data described below.
The recalled data are then both distributionally aligned and information-rich, and a new data recipe requires no fresh classifier~\citep{datologyai2025beyondweb,yu2025repro}.
% duplicated with Data Orchestration / Linear switching): under a linear schedule that shifts their weights during training.

\paragraph{Knowledge-Dense Data Rewriting and Synthesis.}
Rewriting reformulates existing documents, such as encyclopedia articles and specialized STEM material, into information-dense variants: problems, explanations, and knowledge graphs.
A variant keeps its facts fixed and varies only how they are presented, so one fact is learned under several forms, and the formats introduced at this stage are derived from the existing corpus rather than collected anew.
To add long-tail, high-difficulty knowledge without losing factual fidelity, every rewritten or synthesized sample passes a multi-gate filter: automated checks, LLM-as-a-judge evaluation, and manual spot-checking.
For code and mathematics, the sample must also execute in a sandbox or pass a verifier.
Synthesis is bounded to close identified capability gaps, which limits over-reliance and the risk of model collapse~\citep{zhu2024how,gerstgrasser2024is}.

\paragraph{Agentic-Style Data.}
Agentic-style data expose the model to responses that are sequences of tool calls and observations, and come from software-engineering artifacts: issue-pull-request pairs, whole repositories, and commits, with their execution and build logs~\citep{zeng2026davincidev,zhu2026pull}.
Reading a repository as a whole and pairing commits with the issues they resolve let the model learn the sub-skills of this change process---navigating, editing, testing, and revising---rather than finished code alone.
The artifact types cover different parts of the process, and we balance the mixture across them.
To mitigate judge bias, a cascaded filter combines keyword heuristics, model-based scoring, and multi-round verification to screen out reward-hacking and low-quality trajectories~\citep{jiang2026midtool}.

\paragraph{Long-Context Data Curation.}
Long-context training data are built primarily from naturally long documents, which preserve the corpus's domain distribution and discourse structure.
Long-context utility is assessed with \textsc{LongFilter}~\citep{deng2026beyondlength}, which compares the per-token cross-entropy of a complete document with that of its independently scored local chunks. Documents whose continuations are better predicted under long-range context are retained.
Natural length alone provides limited supervision for integrating information across distant positions.
We therefore add a bounded share of synthetic rearrangement and concatenation samples. Their solutions require attending to evidence scattered across the full target context. We also include agentic trajectories, which require state tracking over long action-observation sequences.
These supplements pass the same quality gates as natural documents.
The mixture is specified under three constraints: the category proportions of the base recipe, a global short-to-long ratio, and a stage-dependent preference for longer length buckets.
Iterative proportional fitting solves the three constraints jointly.
Training proceeds in three stages with sequence lengths of 32K, 128K, and 256K, starting from the 8K context used in pre-training.
The token budget is distributed across the stages, with the largest share allocated to the shortest stage, where attention is least expensive.

Ultimately, this extended pre-training phase yields a robust foundation model. It features a distribution aligned to downstream tasks, robust long-range attention capabilities, and preserved trainability, providing a solid and comprehensive basis for the subsequent evaluation and alignment phases.

\subsubsection{Deduplication and Decontamination}

Deduplication and decontamination run continuously as the corpus is assembled.
Deduplication removes exact duplicates via content hashing, near-duplicates via MinHash, and semantic duplicates via domain-adapted embedding models, which protect long-tail knowledge from over-deduplication.
% Decontamination keeps reported scores a measure of capability: leakage turns the benchmark into a test of overlap with the training corpus.
For decontamination,
we index the evaluation sets and compute a continuous overlap score for every training document. Documents above a threshold are removed during final review.
This continuous formulation allows the threshold to be revised without a rescan, mitigating leakage introduced by the collected corpus or by synthetic derivations~\citep{qin2026benchleaking}.
\label{sec:pretrain:data}

% To accommodate both the horizontal and vertical scaling of the model, we design a dedicated data curriculum strategy. Prior work suggests that model scaling should be accompanied by a corresponding shift in the training data distribution\cite{}. Inspired by this insight, we gradually increase data difficulty and quality through a linear mixing schedule as model capacity grows, enabling the expanded model to continue improving on high-quality data. Specifically, we train a classifier to assign each data sample to an appropriate training stage in advance, based on its source and content. During training, the sampling ratios are dynamically adjusted, and this linear transition prevents abrupt shifts in the output distribution, thereby avoiding loss spikes.

\subsection{Data Orchestration}
\label{sec:pretrain:orchestration}
\subsubsection{InfiCoScale}
% As \modelname{} grows in width and depth~(\autoref{sec:pretrain:upscale}), its training data evolve along with it.
% Prior work suggests that model scaling should be accompanied by a corresponding shift in the training data distribution~\citep{singh2026cgls}.
% \emph{InfiCoScale} therefore increases the difficulty and quality of the data step by step as the model capacity grows, so that the grown model keeps improving on high-quality data and the use of ineffective data is reduced.
% Specifically, a classifier assigns each sample to a training stage in advance based on its source and content.
% During training, the sampling ratios of the stages are adjusted with a linear mixing schedule, which avoids abrupt shifts in the output distribution and thus prevents loss spikes.
Humans first master foundational knowledge and only then move on to more complex material. We follow a similar growth path for our model: as \modelbase{} grows in width and depth~(\autoref{sec:pretrain:setup}), its training data also scales in diversity and difficulty along with it.
Prior work suggests that model scaling should be accompanied by a corresponding shift in the training data distribution~\citep{singh2026cgls}.
We therefore introduce the \emph{InfiCoScale} strategy, which scales the training data alongside each model expansion, coordinating architectural growth with progressively higher data quality.
Moreover, we tailor data transitions to the direction of expansion, broadening data coverage when adding experts and increasing data complexity when adding layers.
Specifically, we categorize data sources and form percentile bins from sampled quality and complexity assessments.
During width expansion, we introduce more long-tail and knowledge-rich data. This choice draws on evidence that fine-grained routed experts can specialize~\citep{deepseekmoe} and that adding experts at a fixed active parameter count can improve factual recall~\citep{jelassi2024parrots}.
During depth expansion, we prioritize complex mathematics and code, two domains used to train the newly added layers in LLaMA Pro~\citep{llamapro}. We use these data to target multi-step reasoning, motivated by evidence that deeper models can solve longer reasoning chains on synthetic math problems~\citep{ye2024physics21}.
Each expansion triggers the corresponding data switch, with later stages favoring higher quality bins.
% As \modelbase{} grows in width and depth~(\autoref{sec:pretrain:upscale}), its training data evolve along with it.
% Prior work suggests that model scaling should be accompanied by a corresponding shift in the training data distribution~\citep{singh2026cgls}.
% We therefore introduce the \emph{InfiCoScale} strategy, which pairs each model expansion with a corresponding upshift in training data, coordinating architectural growth with progressively higher data quality.
% Moreover, we tailor data transitions to the direction of expansion, broadening data coverage when adding experts and increasing data complexity when adding layers.
% Specifically, we categorize data sources and form percentile bins from sampled quality and complexity assessments.
% During width expansion, we introduce more long-tail and knowledge-rich data. This choice draws on evidence that fine-grained routed experts can specialize~\citep{deepseekmoe} and that adding experts at a fixed active parameter count can improve factual recall~\citep{jelassi2024parrots}.
% During depth expansion, we prioritize complex mathematics and code, two domains used to train the newly added layers in LLaMA Pro~\citep{llamapro}. We use these data to target multi-step reasoning, motivated by evidence that deeper models can solve longer reasoning chains on synthetic math problems~\citep{ye2024physics21}.
% Each expansion triggers the corresponding data switch, with later stages favoring higher quality bins.

\subsubsection{Linear Switching}
We further apply linear switching to achieve a smooth transition of the data mixture. The key idea is to treat data mixture changes as a gradual process rather than a hard switch. During the constant learning-rate phase, we progressively adjust the sampling distribution.
%: the Chinese-to-English ratio moves from 1:4 to 1:1.8, and code and STEM data are given more weight over time
After the target mixture is reached, it stays unchanged for the rest of pretraining. This smooth transition prevents abrupt distribution shifts, which in turn reduces loss spikes and avoids forgetting earlier knowledge. In practice, this strategy improves overall metrics by about 0.5 points on a 16B MoE model under the same token budget.

\subsubsection{Global Shuffling}
\label{sec:pretrain:orchestration:global-shuffling}
% 公开的大模型预训练实现对 packing 后的序列采用了不同的打乱与混合策略。OLMo 3 32B 在第一阶段对整个混合语料的定长训练序列索引进行全局打乱。SmolLM3 则离线打乱 tokenized sequences，在训练时保留独立的数据源，并按权重交错取样。Megatron 的实现中也采用源内打乱、源间加权混合的方式。我们在 MoE 预训练中比较了阶段内全局打乱与域内打乱后按权重交错两种策略。在较大规模 MoE 配置上测试，全局打乱取得了更低的验证集语言建模损失。我们还观察到，该配置下的序列级专家负载均衡辅助损失更低，单个 microbatch 内各专家接收的 token 数也更均衡。这些结果提示，训练序列的排列方式可能通过改变局部批次的组成，影响专家路由和负载分布。
Public implementations of LLM pretraining adopt different strategies for shuffling and mixing packed sequences. In its first stage, Olmo~3 32B globally shuffles the indices of fixed-length training sequences across the entire data mixture~\citep{olmo3}. SmolLM3 instead shuffles tokenized sequences offline, keeps data sources separate during training, and interleaves samples from them according to their mixture weights~\citep{benallal2025smolplaybook}. Megatron Core similarly shuffles samples within each source and mixes sources by weight~\citep{megatron,nvidia2025megatrondata}.
% In our MoE pretraining, we compare two strategies: global shuffling within each stage, and shuffling within each domain followed by weighted interleaving. On a larger-scale MoE configuration, global shuffling yields a lower validation loss. Under this configuration, we also observe a lower sequence-level auxiliary loss for expert load balancing, and a more even distribution of tokens across experts within each micro-batch. These results suggest that the ordering of training sequences may affect expert routing and load distribution by changing the composition of local batches.
In our MoE pretraining, we compare two strategies: global shuffling within each stage, and shuffling within each domain followed by weighted interleaving. On a larger-scale MoE configuration, global shuffling yields a lower validation loss. It also gives a lower sequence-level auxiliary loss for expert load balancing and a more even distribution of tokens across experts within each micro-batch, which reduces the straggler effect in expert-parallel training. We therefore use global shuffling throughout pre-training.

\subsection{Pre-Training Setups}
\label{sec:pretrain:setup}

\begin{figure}[!h]
\centering
\includegraphics[width=\textwidth]{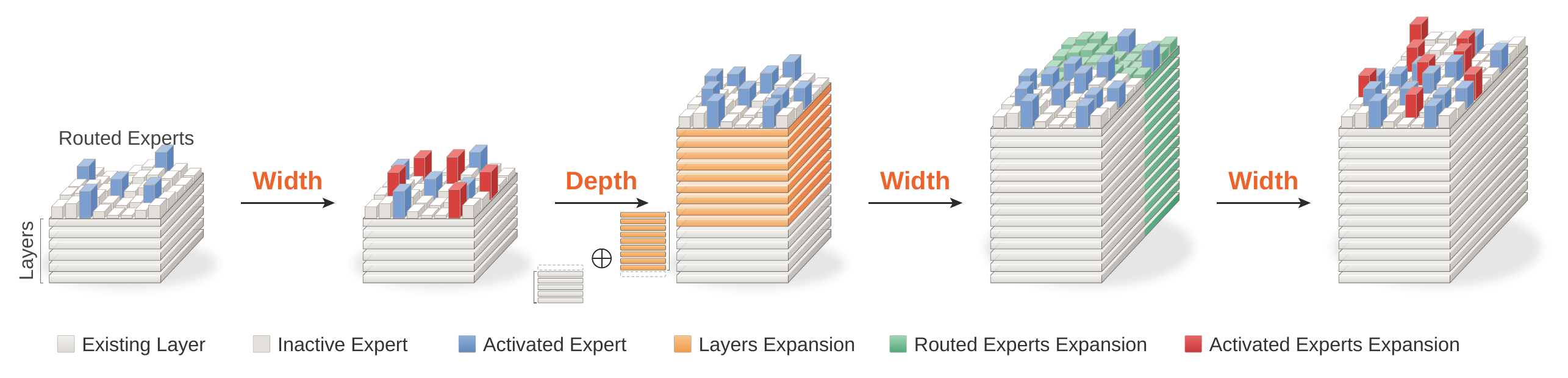}
\caption{Growth process of \modelname{} with InfiGrow during pre-training. Each block is the stack of layers, and the pillars on top are the routed experts of a layer, where the blue pillars are the activated experts. Colors mark the part added at each growth step: more activated experts (red), more layers by stacking the lower half of one copy of the model with the upper half of another copy (orange), and new experts copied from hot experts with noise (green).}
\label{fig:upscale_process}
\end{figure}

\subsubsection{Training Setups}
We initialize the model with a standard deviation of 0.006 and first train it with the AdamW optimizer~\citep{adamw}, using $\beta_1 = 0.9$, $\beta_2 = 0.95$, $\epsilon = 10^{-15}$ and a weight decay of 0.1. During training, the optimizer is then switched to Muon~\citep{muon,moonlight} with a momentum of 0.95, while the embedding and the output head remain optimized with AdamW.
The learning rate is linearly warmed up to $3.5 \times 10^{-4}$ over the first 2,000 steps and kept constant for most of pre-training.
It is then decayed in two short steps, first to $3.0 \times 10^{-4}$ and later to $2.7 \times 10^{-4}$, and finally decayed to $2 \times 10^{-5}$ at the end of pre-training.
In total, \modelname{} is pre-trained on 13.36T tokens with a sequence length of 8K, and the global batch size is gradually increased to about 100M tokens during training (\autoref{sec:pretrain:stability:rampup}).
For load balance, \modelname{} adopts the auxiliary-loss-free strategy~\citep{lossfree,deepseekv3}, with the bias update rate set to $10^{-3}$ and set to 0 at the end of pre-training, while the sequence-wise balance loss coefficient is kept at $10^{-4}$ throughout training.

After pre-training, the context length is extended from 8K to 32K, 128K and 256K in three stages, all using Muon with a constant learning rate of $2 \times 10^{-5}$, and \modelname{} further supports a context length of up to 1M tokens.
During the main pre-training, both FA and SWA layers use a RoPE base frequency of 50,000.
In the long-context training, the RoPE base of FA layers is increased to 10,000,000, while that of SWA layers is kept at 50,000. 

\subsubsection{Training Stability}
\label{sec:pretrain:stability}

Training stability is a major challenge for large-scale MoE models~\citep{switch,stmoe,deepseekv4}.
In \modelname{}, most loss spikes are traced to insufficient numerical precision and activation outliers, and three techniques are deployed to mitigate the instability.
With these techniques, the pre-training of \modelname{} is very stable and never requires rolling back to an earlier checkpoint or reducing the learning rate.

\paragraph{Batch size ramp-up.}
\label{sec:pretrain:stability:rampup}
\modelname{} adopts a dynamic batch size ramp-up~\citep{gpt3,palm}.
A small batch size is more token-efficient early in training~\citep{mccandlish2018,merrill2025cbs}, while progressively increasing the batch size reduces the gradient noise and helps stabilize training.
Compared with using a large batch size from the start, this schedule reaches a similar loss with fewer training tokens and less compute~\citep{merrill2025cbs}.
The global batch size is gradually increased during training, reaching about 100M tokens.

\paragraph{FP32 router, RoPE and expert token counts.}
\label{sec:pretrain:stability:fp32}
Although most of the computation is performed in BF16, several small but precision-sensitive components of \modelname{} are kept in FP32.
\modelname{} finds that the precision of RoPE matters especially in early training, where computing the rotary position embeddings in BF16 leads to noticeably more loss spikes.
The routing path is similarly sensitive: BF16 rounding can make close routing scores indistinguishable and flip the top-$K$ expert selection, and since BF16 cannot exactly represent large integers, the per-expert token counts that drive the load-balance bias update accumulate large rounding errors.
Therefore, the RoPE computation, the router weights and scores, and the per-expert token counts are all computed in FP32, at a negligible extra cost.

\paragraph{SwiGLU clip.}
\label{sec:pretrain:stability:clip}
SwiGLU clipping is not used from the beginning, but is introduced during training after obvious loss spikes are observed.
These spikes coincide with activation outliers in the shared expert of the last layer and in the routed experts of the last three layers, while the 99th percentile of the activations in these experts stays below 10.
Following gpt-oss~\citep{gptoss}, the outputs of the gate and up projections of these experts are clamped to 16, well above the 99th percentile, so that only the outliers are affected.
After clipping is applied, the loss first rises briefly, then drops rapidly and eventually falls below that of the run without clipping.

\subsubsection{Growth Schedule}
\label{sec:pretrain:upscale}

\modelname{} is grown four times during pre-training~(\autoref{fig:upscale_process}), and each growth step affects the training loss differently.
The first width growth only increases the number of activated experts.
Since no new parameters are introduced, this change is mild: the loss is slightly higher than that of the model without growth in the first few steps, but soon becomes lower.
The depth growth is then performed, increasing the number of layers by stacking.
Unlike the first width growth, it introduces a large number of new parameters, so the learning rate is warmed up again over 100 steps.
Although the loss increases obviously right after the depth growth, it converges very fast and drops clearly below the level before growth even before the warm-up finishes.
Then, the second width growth is performed, expanding the routed experts by duplicating the hot experts with noise.
The learning rate is again warmed up over 100 steps.
In contrast to the depth growth, the loss recovers much more slowly after the obvious increase and takes much longer to fall below the level before growth. Finally, \modelname{} increases the number of activated experts again, and the loss shares the same behavior as the first width growth.

\subsection{Evaluations}
\label{sec:pretrain:eval}
% 为了全面对比我们base模型和其他模型的能力，我们从四个领域出发评测了广泛的基准测试，包括语言和推理、知识、代码和数学。

% We evaluate our base model against other models on a broad set of benchmarks covering four domains: language and reasoning, knowledge, code, and math.
Our base model is compared against seven open-source models across a broad set of benchmarks spanning language and reasoning, knowledge, math, and code.

\subsubsection{Evaluation Benchmarks}
The evaluation suite consists of 23 benchmarks in English and Chinese, grouped into the following four domains.

\paragraph{Language \& Reasoning.}
We evaluate general language understanding and reasoning on BBH~\citep{bbh}, HellaSwag~\citep{hellaswag}, WinoGrande~\citep{winogrande}, DROP~\citep{drop}, and SQuAD~\citep{squad} in English, and on C3~\citep{c3} in Chinese.

\paragraph{Knowledge.}
For English knowledge, we use MMLU-Redux~\citep{mmluredux}, MMLU-Pro~\citep{mmlupro}, GPQA-Diamond~\citep{gpqa}, SuperGPQA~\citep{supergpqa}, TriviaQA~\citep{triviaqa}, and ARC-Challenge~\citep{arc}; for Chinese knowledge, we use C-Eval~\citep{ceval} and CMMLU~\citep{cmmlu}. We also report AGIEval~\citep{agieval}, averaged over its English and Chinese subsets.

\paragraph{Math.}
We evaluate English mathematical reasoning on GSM8K~\citep{gsm8k} and MATH~\citep{math}, and Chinese mathematical reasoning on Ape210K~\citep{ape210k} and Math23K~\citep{wang-etal-2017-deep}.

\paragraph{Code.}
We evaluate coding ability on CRUXEval~\citep{cruxeval}, averaged over its input- and output-prediction tasks (CRUXEval-I and CRUXEval-O), HumanEval+ and MBPP+~\citep{evalplus}, and LiveCodeBench~\citep{livecodebench}.

\subsubsection{Benchmark Selection}
\label{sec:bench-selection}

Benchmark suites are usually inherited by convention. Popularity is a reasonable criterion, since a benchmark that nobody else reports cannot be compared against published numbers, but on its own it leaves the measurement uncertainty unquantified~\citep{evalvariance} and accumulates benchmarks that are small, saturated, or near-duplicates of each other. We therefore keep popularity as the primary criterion and gate it on measurement quality, following FineWeb2~\citep{fineweb2} for monotone behavior along the training curve, \citet{evalvariance} and \citet{signalnoise} for direct noise measurement, Arena-Hard~\citep{arenahard} for separability, and Olmo~3~\citep{olmo3} for cross-model correlation as a redundancy test. Starting from a pool of 44 evaluation units, we score each one on \emph{popularity} $p$, the number of the 24 surveyed base model reports that report it; \emph{stability}, the per-question flip noise between end-of-training checkpoints propagated into the score, which also accounts for most non-monotone fluctuations along the training curve; \emph{confidence}, measured by Spearman--Brown-corrected split-half reliability~\citep{spearman1910,brown1910} and by separability $Q$, the fraction of model pairs whose paired-bootstrap difference intervals exclude zero; and \emph{redundancy} $\rho$, the within-domain cross-model rank correlation with the other benchmarks. Two further quantities characterize a whole suite rather than a single benchmark: its signal $\sigma_{\text{models}}$, the spread of the model averages it induces, and its rank agreement with the full candidate pool. We verify that each metric behaves as intended before applying it.

We keep the most-reported benchmarks unless redundant, keep one member of each near-duplicate group ($\rho > 0.6$), and drop benchmarks that are both rarely reported and of low measured quality, yielding the 23 benchmarks in \autoref{tab:base-eval}. Benchmarks left out include MMLU, HumanEval and MBPP, each superseded by a cleaner variant we keep.
Four are kept \emph{despite} falling below the quality floor, as omitting them would make our numbers incomparable: GPQA-Diamond, LiveCodeBench, HumanEval+ and ARC-Challenge are the rows on which small differences should not be over-interpreted. \autoref{tab:bench-selection} compares the resulting suite with random draws of the same per-domain size and with a suite chosen by popularity alone. Ours improves on the mean of the random draws on every criterion except rank agreement, where unbiased random draws match the unfiltered pool in expectation and ours stays well within their spread. By contrast, the popularity-only suite attains higher popularity by construction but is worse on every quality axis, as it retains all near-duplicate families.

% Generated from work/scripts/selection_experiments.py
\begin{table}[ht!]
\centering
\small
\caption[Benchmark suite comparison]{
    Our benchmark suite compared with alternative selection strategies, all restricted to the same candidate pool of 44 evaluation units and to the same number of benchmarks per domain.
    ``Random'' summarizes 5{,}000 random draws and ``Popularity-only'' takes the most frequently reported benchmarks per domain.
    Arrows indicate the preferred direction; the best value in each row is in bold.
}
\label{tab:bench-selection}
\setlength{\tabcolsep}{6pt}
\renewcommand{\arraystretch}{1.1}
\begin{tabular}{lccc}
\toprule
\textbf{Criterion} & \textbf{Ours} & \textbf{Random} \textit{mean [p5, p95]} & \textbf{Popularity-only} \\
\midrule
Popularity $\bar{p}$ $\uparrow$ & 11.0 & 9.1 [7.7, 10.5] & \textbf{12.7} \\
Separability $\bar{Q}$ (\%) $\uparrow$ & \textbf{74.6} & 71.8 [67.7, 75.8] & 71.9 \\
Split-half reliability $\uparrow$ & \textbf{0.883} & 0.866 [0.842, 0.891] & 0.867 \\
Redundancy $\bar{\rho}$ $\downarrow$ & \textbf{0.349} & 0.376 [0.337, 0.417] & 0.409 \\
Signal $\sigma_{\text{models}}$ $\uparrow$ & \textbf{1.82} & 1.69 [1.42, 1.95] & 1.56 \\
Rank agreement $\tau$ with full pool & 0.769 & 0.782 [0.615, 0.923] & 0.667 \\
\bottomrule
\end{tabular}
\end{table}

\subsubsection{Evaluation Settings}
\paragraph{Comparison Models.}
We compare \modelbase{} with several recent open-source base models:
DeepSeek-V4-Flash~\citep{deepseekv4}, DeepSeek-V4-Pro~\citep{deepseekv4},
MiMo-V2.5~\citep{mimov25}, Hy3-preview~\citep{hy3},
Ling-3.1-Flash~\citep{ling31flash}, Mistral-Large-3~\citep{mistrallarge3},
and Step-3.5-Flash~\citep{step35}. All of them are MoE models that are widely adopted by the community, with 196B--1.57T total and 10.9B--47.9B activated parameters, covering and exceeding the scale of \modelbase{} (209.6B total, 14.4B activated). For all models, parameter counts exclude the embedding layers and the multi-token prediction (MTP) module. %我们比较的均为相似参数量级的MoE base模型，这些模型均发布于2026年且获得了社区的认可。

\paragraph{Evaluation Details.}
All benchmark evaluations are conducted on our internal evaluation framework, with a unified configuration across all models to ensure a fair comparison. We evaluate base checkpoints directly using greedy decoding. 
Most benchmarks are evaluated generatively and scored via rule-based answer extraction; 
only HellaSwag is evaluated without generation using the per-token mean log-probability of each candidate answer.
Code benchmarks are evaluated by executing generated programs against test suites (pass@1). 
\autoref{app:benchmark_prompt} explains the prompt design and representative prompts.

\subsubsection{Evaluation Results}

\definecolor{zhrow}{RGB}{240,247,254}  % light blue background for Chinese benchmarks
% Generated by work/scripts/gen_base_eval_table_v3.py from work/resources/infix 预训练评测结果.xlsx (benchmark set V3)
\begin{table}[ht!]
\centering
\small
\caption[Base model comparison]{
    Comparison of \textbf{\modelbase{}} with open-source base models. Chinese benchmarks are shaded in blue. The best result in each row is in bold and the second best is underlined. $^\dagger$AGIEval is the average of its English and Chinese subsets; $^\ddagger$CRUXEval is the average of CRUXEval-I and CRUXEval-O.% The average is computed over all 23 benchmarks.
}
\label{tab:base-eval}
\setlength{\tabcolsep}{3.5pt}
\renewcommand{\arraystretch}{1.3}
\resizebox{\textwidth}{!}{%
\begin{tabular}{l>{\footnotesize}r|c|ccccccc}
\toprule
\multirow{2}{*}{\textbf{Benchmark}} & \multirow{2}{*}{\textbf{Metric}} & \begin{tabular}[c]{@{}c@{}}\textbf{\modelname{}}\\\textbf{Base}\end{tabular} & \begin{tabular}[c]{@{}c@{}}\textbf{DeepSeek}\\\textbf{V4-Flash}\end{tabular} & \begin{tabular}[c]{@{}c@{}}\textbf{DeepSeek}\\\textbf{V4-Pro}\end{tabular} & \begin{tabular}[c]{@{}c@{}}\textbf{MiMo}\\\textbf{V2.5}\end{tabular} & \begin{tabular}[c]{@{}c@{}}\textbf{Hy3}\\\textbf{preview}\end{tabular} & \begin{tabular}[c]{@{}c@{}}\textbf{Ling-3.1}\\\textbf{Flash}\end{tabular} & \begin{tabular}[c]{@{}c@{}}\textbf{Mistral}\\\textbf{Large-3}\end{tabular} & \begin{tabular}[c]{@{}c@{}}\textbf{Step-3.5}\\\textbf{Flash}\end{tabular} \\
 & & {\scriptsize\begin{tabular}[c]{@{}c@{}}\textit{209.6B}\\\textit{A14.4B}\end{tabular}} & {\scriptsize\begin{tabular}[c]{@{}c@{}}\textit{283.3B}\\\textit{A12.7B}\end{tabular}} & {\scriptsize\begin{tabular}[c]{@{}c@{}}\textit{1571.2B}\\\textit{A47.9B}\end{tabular}} & {\scriptsize\begin{tabular}[c]{@{}c@{}}\textit{308.5B}\\\textit{A15.2B}\end{tabular}} & {\scriptsize\begin{tabular}[c]{@{}c@{}}\textit{294B}\\\textit{A19.7B}\end{tabular}} & {\scriptsize\begin{tabular}[c]{@{}c@{}}\textit{560B}\\\textit{A25B}\end{tabular}} & {\scriptsize\begin{tabular}[c]{@{}c@{}}\textit{674.1B}\\\textit{A40.6B}\end{tabular}} & {\scriptsize\begin{tabular}[c]{@{}c@{}}\textit{195.9B}\\\textit{A10.9B}\end{tabular}} \\
\midrule

\rowcolor{black!5}
\multicolumn{10}{l}{\large\textit{Language \& Reasoning}} \\
BBH & \textit{3-shot CoT, EM} & 88.07 & 88.73 & \textbf{90.25} & \underline{89.94} & 88.36 & 89.79 & 88.88 & 87.91 \\
HellaSwag & \textit{10-shot, PPL} & 86.53 & 84.66 & \underline{87.18} & 85.38 & 85.53 & 82.71 & 84.91 & \textbf{88.62} \\
WinoGrande & \textit{3-shot, EM} & 79.01 & \underline{79.72} & \textbf{81.77} & 78.93 & 78.85 & 73.24 & 65.51 & 67.88 \\
DROP & \textit{5-shot CoT, EM} & 83.50 & 84.90 & \underline{84.94} & 83.76 & 82.49 & \textbf{85.74} & 82.69 & 80.68 \\
SQuAD & \textit{3-shot, EM} & \underline{82.41} & 76.56 & 80.40 & 74.02 & 74.25 & \textbf{85.66} & 75.05 & 75.48 \\
\rowcolor{zhrow}
C3 & \textit{5-shot, EM} & 97.20 & 97.69 & \textbf{98.23} & 97.02 & 97.56 & 70.76 & \underline{97.74} & 97.15 \\
\midrule

\rowcolor{black!5}
\multicolumn{10}{l}{\large\textit{Knowledge}} \\
MMLU-Redux & \textit{5-shot, EM} & 91.13 & 91.67 & \textbf{93.00} & 90.88 & 90.44 & \underline{92.51} & 91.32 & 90.46 \\
MMLU-Pro & \textit{5-shot CoT, EM} & 69.37 & 65.98 & 67.92 & 66.45 & \underline{70.92} & \textbf{73.54} & 68.09 & 61.68 \\
GPQA-Diamond & \textit{5-shot CoT, EM} & 60.61 & 56.57 & \underline{64.14} & \textbf{67.68} & 60.10 & 63.57 & 59.09 & 54.04 \\
SuperGPQA & \textit{5-shot CoT, EM} & 43.69 & 43.00 & \underline{44.09} & 40.02 & 43.00 & \textbf{45.30} & 43.97 & 39.83 \\
TriviaQA & \textit{5-shot, EM} & 84.48 & \underline{86.60} & \textbf{89.03} & 84.55 & 85.81 & 86.15 & 72.48 & 85.50 \\
ARC-Challenge & \textit{25-shot, EM} & 95.65 & 95.99 & 97.01 & 96.59 & 96.42 & \textbf{97.87} & \underline{97.44} & 95.39 \\
AGIEval$^\dagger$ & \textit{5-shot, EM} & 83.32 & \underline{83.57} & \textbf{84.50} & 78.49 & 81.20 & 83.26 & 75.69 & 77.68 \\
\rowcolor{zhrow}
C-Eval & \textit{5-shot, EM} & \textbf{93.31} & 92.05 & \underline{92.87} & 87.59 & 90.86 & 88.56 & 88.41 & 89.38 \\
\rowcolor{zhrow}
CMMLU & \textit{5-shot, EM} & \textbf{91.21} & 90.51 & \underline{91.17} & 87.64 & 90.01 & 80.06 & 86.84 & 89.57 \\
\midrule

\rowcolor{black!5}
\multicolumn{10}{l}{\large\textit{Math}} \\
GSM8K & \textit{4-shot CoT, EM} & \textbf{96.29} & 95.68 & \underline{96.13} & 95.15 & \textbf{96.29} & \textbf{96.29} & 95.83 & 95.07 \\
MATH & \textit{4-shot CoT, EM} & 64.10 & 62.02 & \underline{68.16} & 63.62 & 65.80 & \textbf{74.80} & 61.94 & 62.74 \\
\rowcolor{zhrow}
Ape210K & \textit{3-shot CoT, EM} & \textbf{83.84} & 81.78 & 80.82 & 75.92 & \underline{83.60} & 81.76 & 73.34 & 80.58 \\
\rowcolor{zhrow}
Math23K & \textit{3-shot CoT, EM} & \underline{95.20} & 95.10 & 94.50 & 91.50 & 94.20 & \textbf{95.70} & 87.70 & 87.30 \\
\midrule

\rowcolor{black!5}
\multicolumn{10}{l}{\large\textit{Code}} \\
CRUXEval$^\ddagger$ & \textit{2-shot CoT, Pass@1} & 80.07 & 77.62 & 78.19 & 78.06 & \underline{80.25} & \textbf{89.57} & 72.50 & 71.06 \\
HumanEval+ & \textit{3-shot, Pass@1} & 84.76 & 84.15 & 82.32 & \underline{87.20} & 83.54 & \textbf{89.63} & 81.10 & 82.32 \\
MBPP+ & \textit{3-shot, Pass@1} & \underline{77.78} & \underline{77.78} & 77.51 & 75.93 & 74.87 & \textbf{78.57} & 76.19 & 74.60 \\
LiveCodeBench & \textit{2-shot, Pass@1} & \textbf{41.98} & 31.30 & 36.64 & 31.30 & 32.82 & \underline{40.46} & 25.19 & 33.59 \\
\midrule
\textbf{Average} &  & \underline{80.59} & 79.29 & \textbf{80.90} & 78.59 & 79.44 & 80.24 & 76.17 & 76.89 \\
\bottomrule
\end{tabular}%
}
\end{table}

\paragraph{Overall.}
As shown in \autoref{tab:base-eval}, \modelbase{} achieves an average score of 80.59 over the 23 benchmarks, on par with DeepSeek-V4-Pro (80.90), and ahead of all other compared models, including Ling-3.1-Flash (80.24), Hy3-preview (79.44), and DeepSeek-V4-Flash (79.29). Notably, \modelbase{} reaches this level with only 209.6B total and 14.4B activated parameters, about 1/7 of the total and less than 1/3 of the activated parameters of DeepSeek-V4-Pro (1.57T total, 47.9B activated).

\paragraph{Language \& Reasoning.}
In general language understanding and reasoning, \modelbase{} ranks second among all models by domain average (86.12), on par with DeepSeek-V4-Pro and ahead of DeepSeek-V4-Flash by 0.74 points. It achieves the second-best result on SQuAD (82.41), outperforming DeepSeek-V4-Pro and DeepSeek-V4-Flash by 2.01 and 5.85 points, respectively, and ranks among the top three on HellaSwag and WinoGrande.

\paragraph{Knowledge.}
\modelbase{} demonstrates strong knowledge capabilities, particularly in Chinese, ranking second by domain average (79.20) and on par with DeepSeek-V4-Pro. It achieves the best results on C-Eval (93.31) and CMMLU (91.21), ahead of all compared models, including DeepSeek-V4-Pro. On challenging English knowledge benchmarks, it ranks among the top three on MMLU-Pro (69.37) and AGIEval (83.32), and outperforms DeepSeek-V4-Flash on MMLU-Pro (+3.39), GPQA-Diamond (+4.04), and SuperGPQA (+0.69).

\paragraph{Math.}
In mathematical reasoning, \modelbase{} achieves the best results on GSM8K (96.29, tied with Hy3-preview and Ling-3.1-Flash) and Ape210K (83.84), and the second-best on Math23K (95.20). On Ape210K, it outperforms DeepSeek-V4-Pro and DeepSeek-V4-Flash by 3.02 and 2.06 points, respectively. On the more challenging competition-level MATH benchmark, it scores 64.10, ahead of DeepSeek-V4-Flash (+2.08), MiMo-V2.5, Mistral-Large-3, and Step-3.5-Flash.

\paragraph{Code.}
\modelbase{} exhibits a prominent advantage in code, ranking second by domain average (71.15) and outperforming DeepSeek-V4-Pro by 2.49 points. It achieves the best result on LiveCodeBench (41.98), outperforming DeepSeek-V4-Pro by 5.34 points and DeepSeek-V4-Flash by 10.68 points, and the second best on MBPP+ (77.78). It also outperforms both DeepSeek-V4-Pro and DeepSeek-V4-Flash on CRUXEval and HumanEval+.

%%======================================================================
\subsection{Intelligence per Compute}
\label{sec:pretrain:ipc}
\begin{figure}[!t]
\centering
\includegraphics[width=0.65\textwidth]{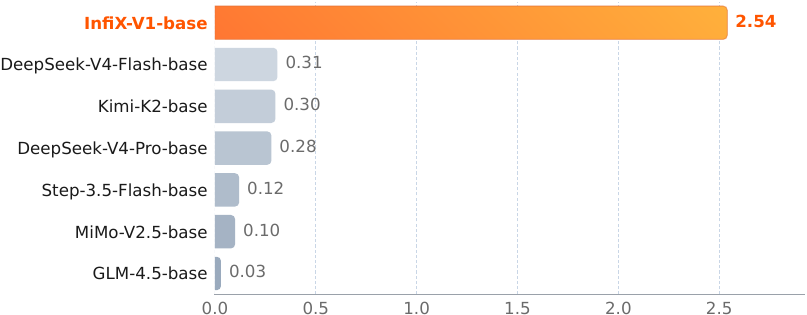}
\caption{Intelligence per compute~(IPC, \autoref{eq:intelligence_per_compute}) of \modelbase{} and the compared base models, computed with $\alpha$ fitted on the open baseline models.}
\label{fig:ipc}
\end{figure}
Besides the final capability, we also measure how much capability a model gains from its training compute.
Existing metrics each capture only part of this.
MFU~\citep{palm} measures how well the hardware is used, but not what the model learns from it.
Intelligence indices~\citep{artificialanalysis2025} compare the final capability of models, but ignore the cost of reaching it.
Intelligence per watt~\citep{saadfalcon2025ipw} relates capability to energy, but only at inference time.
We therefore define \emph{intelligence per compute}~(IPC) for pre-training.
It builds on the assumption that the closer a model gets to a perfect score, the more compute each additional point costs.
Following neural scaling laws~\citep{hestness2017deep,kaplan2020scaling} and the increasing marginal cost in economics, we let the marginal cost of raising the average benchmark score $S \in [0,1]$ grow with the compute $C$ already spent and with the inverse of the error rate $1-S$:
\begin{equation}
\frac{\mathrm{d}C}{\mathrm{d}S} = \frac{C}{\alpha\,(1-S)}
\label{eq:marginal_cost}
\end{equation}
where $\alpha$ controls how fast the cost grows.
Integrating \autoref{eq:marginal_cost} gives $C = K(1-S)^{-1/\alpha}$.
Here $K$ depends only on the training recipe, and a smaller $K$ means that the recipe needs less compute to reach any score.
We take its inverse as IPC:
\begin{equation}
\eta = \frac{1}{K} = \frac{(1-S)^{-1/\alpha}}{C}.
\label{eq:intelligence_per_compute}
\end{equation}
To compare IPC, $\alpha$ is obtained by fitting $\ln(1-S) = c - \alpha \ln C$ on the open baseline models in \autoref{fig:teaser}.
With the fitted $\alpha$, \modelbase{} achieves the highest $\eta$ among all compared models, about 8 times that of the second best~(\autoref{fig:ipc}).

\section{Post-Training}
\label{sec:posttrain}

\begin{figure}[!htbp]
\centering
\captionsetup{font=small,skip=6pt}
% Exported from posttraining_pipeline.drawio; the matching SVG is retained in figures/.
\includegraphics[width=\textwidth,height=0.42\textheight,keepaspectratio]{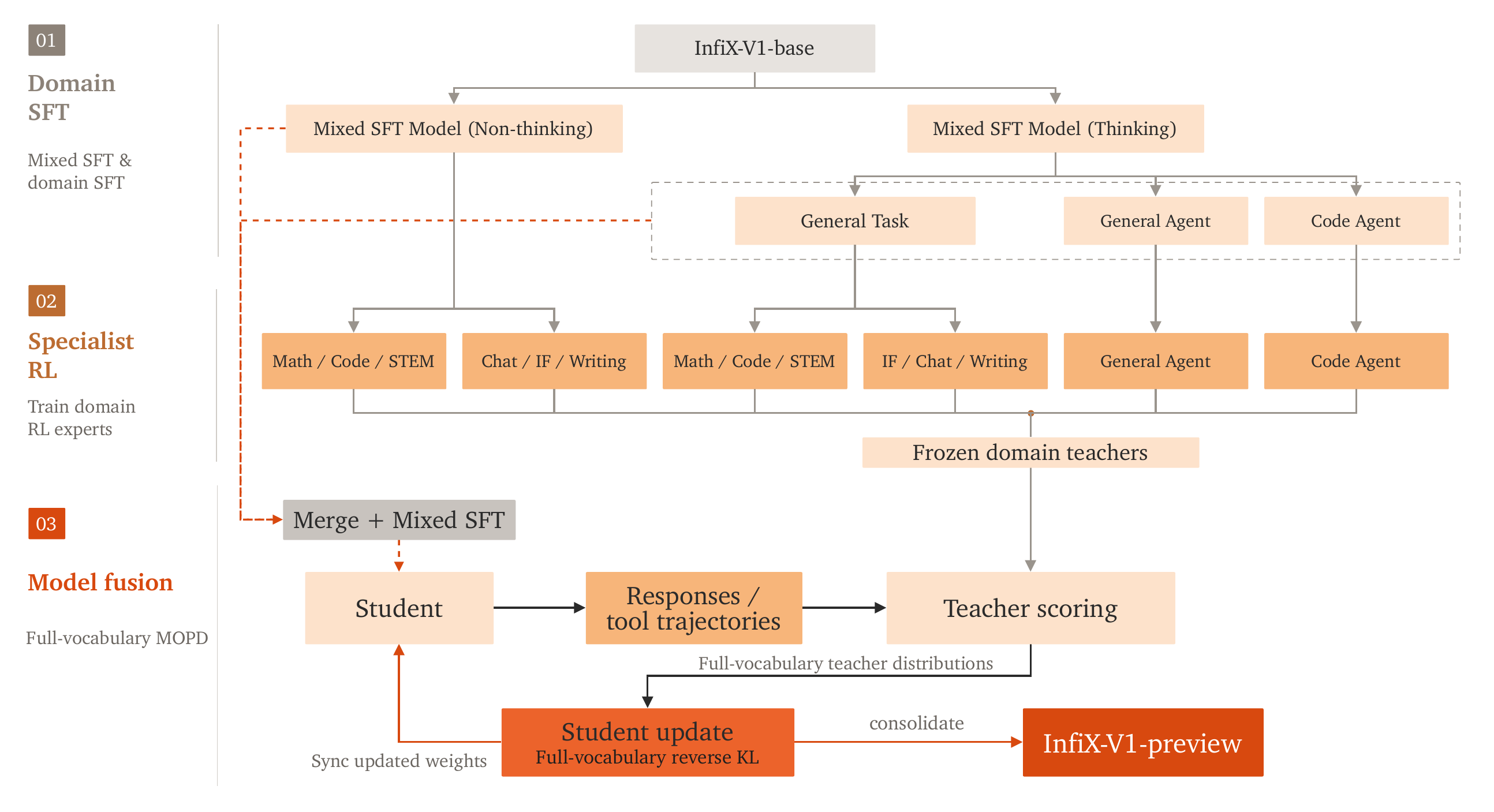}
\caption{Post-training pipeline of \modelname{}.}
\label{fig:posttraining_pipeline}
\end{figure}

The post-training of \modelname{} consists of three stages, as illustrated in \autoref{fig:posttraining_pipeline}: \textbf{(1) Supervised fine-tuning} builds mixed and domain SFT models from \modelbase{} (\autoref{sec:posttrain:specialist_sft}); \textbf{(2) Specialist RL} trains domain RL specialists that serve as teachers (\autoref{sec:posttrain:specialist_rl}); and \textbf{(3) Model fusion} merges the SFT models into a student and distills all specialists into a single model via full-vocabulary MOPD (\autoref{sec:posttrain:fusion}).

\subsection{Supervised Fine-Tuning}
\label{sec:posttrain:specialist_sft}

We build a broad multi-domain corpus from public datasets and synthetic trajectories, and verify it in multiple stages with rule-based filters, execution checks, and model graders for correctness and consistency. Supervised fine-tuning of \modelbase{} on this corpus yields two variants that differ in thinking behavior: a non-thinking model, which answers directly, and a thinking model, which emits an explicit reasoning trace before answering. The thinking model is then specialized, on dedicated mixtures, into three domain models: General Task, General Agent, and Code Agent. The General Task mixture covers general chat and writing, instruction following and alignment, mathematics, STEM, and code generation. The General Agent mixture covers web search, deep research, function calling, and multi-turn tool use. The Code Agent mixture covers repository-level software engineering and terminal tasks.

% Verify SFT domain coverage, trace formatting, and target masking against the final runs.

\subsection{Draft Model Training}
\label{sec:posttrain:dflash}
Long-horizon RL rollouts generate extensive token sequences, 
making autoregressive decoding a primary bottleneck in both training and evaluation. 
To accelerate generation without altering the target model's outputs, we train the draft model (\autoref{sec:arch:dflash}) for speculative decoding.
Since speculative verification accepts only the prefix before the first rejection, 
fixed positional decay cannot adapt to the positions that bottleneck acceptance across examples or training stages. 
We therefore adopt the D-PACE loss~\citep{dpace}, which dynamically weights each position by its contribution to the expected accepted length. 
For a block with $m$ supervised positions, the objective is
\begin{equation}
\mathcal{L}_{\mathrm{D-PACE}} = \sum_{i=1}^{m} w_i (-\log q_i),
\label{eq:posttrain:dpace}
\end{equation}
where $q_i$ is the probability assigned by the draft model to the reference token at position $i$, and $w_i$ is a dynamic position weight derived from smoothed prefix confidences and treated as constant during backpropagation.
We train the draft model on target-generated rollouts for 60{,}000 steps with a block size of 8 and up
to 512 anchors per sequence. Under speculative rejection sampling, the resulting draft model achieves a mean acceptance length of 4.26 tokens per step on a mixed-domain evaluation set. 
In practice, this translates into consistent end-to-end speedups over standard decoding, reaching $1.96\times$ on agentic coding tasks and substantially shortening both training and evaluation cycles.

\subsection{Domain-Specialized Reinforcement Learning}
\label{sec:posttrain:specialist_rl}

Starting from the SFT checkpoints, we perform domain-specialized RL across both response modes~\citep{deepseekv4}, split into general RL for text-only tasks (mathematics, code generation, STEM, instruction following, chat, and writing) and agentic RL for multi-turn tool use in interactive environments.
Each specialist is optimized separately on its target domain and later serves as a frozen teacher during model fusion.

% Verify production reward rules, the RL optimizer, and task mixtures against run configs.
\paragraph{General RL.}
Reward signals follow task verifiability. For tasks with verifiable outcomes, we apply Reinforcement Learning with Verifiable Rewards (RLVR): math is scored by symbolic or numerical equivalence, logic and instruction following by constraint satisfaction, and coding by unit-test execution in a sandbox. Tasks requiring semantic judgment, such as open-ended dialogue and writing, are scored by a pairwise LLM judge conditioned on a reference response~\citep{deepseekv4}.

\paragraph{Agentic RL.}
We conduct domain-specific agentic RL independently across Terminal~\citep{terminal_bench}, SWE~\citep{deepswe,swebench_verified,swebench_pro}, and general-agent tasks~\citep{browsecomp,browsecomp_zh,tau3bench}. Our training infrastructure supports multiple agent harnesses with task-specific environments and verification procedures. Agents interact with isolated sandboxes to complete multi-turn tasks, while domain-specific verifiers and rollout graders provide correctness and quality signals for training.

\textit{Concurrent rollout execution.}
Rollouts are collected in task-level groups using a pool of concurrent sandboxes. Whenever a group slot becomes available, we dispatch a new task with its corresponding group of rollouts for the same task to sustain utilization. Sandbox concurrency is configured separately from the inference engine's maximum running requests, since each agent alternates between model generation and tool execution.

\textit{Training--inference consistency.}
R3~\citep{ma2025r3} replays recorded expert selections during training, while token-in-token-out preserves original token IDs across multi-turn interactions and training.

\textit{Staged correctness rewards.}
We adopt a two-stage correctness optimization. For repository-based tasks with decomposed test labels, the first stage rewards partial repair progress while requiring all regression tests to remain satisfied:
\begin{equation}
    r_i^{(1)}
    = f_i^{\mathrm{F2P}}\,
      \mathbb{I}\!\left[f_i^{\mathrm{P2P}}=1\right],
    \label{eq:posttrain:coding_agent_stage1}
\end{equation}
where $f_i^{\mathrm{F2P}}$ and $f_i^{\mathrm{P2P}}$ are the fractions of F2P and P2P tests passed. The second stage switches to a strict binary correctness signal:
\begin{equation}
    r_i^{(2)}
    = \mathbb{I}\!\left[
        f_i^{\mathrm{F2P}}=1
        \;\land\;
        f_i^{\mathrm{P2P}}=1
      \right].
    \label{eq:posttrain:coding_agent_stage2}
\end{equation}
The first stage supplies credit for incremental fixes, while the second targets complete task resolution. Tasks without an F2P/P2P decomposition use their native verification signal.

Following MiMo-V2.6~\citep{mimo2026v26}, we complement test-based rewards with rollout grading and behavioral regularization. Groupwise reward synthesis (GRS) multiplies the test reward $r_i^{\mathrm{test}}$ by task-specific solution and behavior scores $s_i^{\mathrm{sol}}, s_i^{\mathrm{beh}}\in[0,1]$ derived from offline-rollout rubrics, while confirmed reward hacking receives zero reward ($h_i=0$, and $h_i=1$ otherwise):
\begin{equation}
    R_i^{\mathrm{GRS}}
    = h_i\, r_i^{\mathrm{test}}\, s_i^{\mathrm{sol}}\, s_i^{\mathrm{beh}}.
    \label{eq:posttrain:coding_agent_grs}
\end{equation}
The integrity gate $h_i$ covers three categories of reward hacking: access to restricted answers or reference solutions; unauthorized online searches or downloads; and circumvention or manipulation of test-based evaluation, including test tampering and fabricated grading results. We set $h_i=0$ only when such violations are substantiated by rollout evidence.

Groupwise advantage redistribution (GAR) instead ranks passing trajectories within mixed-outcome groups and reallocates their positive advantages toward higher-quality solutions, with rescaling and recentering to control the group-level learning signal~\citep{dong2026gagar}. Given binary outcomes $r_i = h_i r_i^{\mathrm{test}}\in\{0,1\}$ in a group of size $G$, the passing set $\mathcal{P}=\{i:r_i=1\}$ ($0<|\mathcal{P}|<G$), baseline advantages $a_i = r_i - \frac{1}{G}\sum_{j=1}^G r_j$, rank weights $w_i\in(0,1]$, and cap $\lambda_{\max}>1$,
\begin{equation}
    \lambda = \min\!\left(\lambda_{\max},\, \frac{\sum_{j\in\mathcal{P}} a_j}{\sum_{j\in\mathcal{P}} w_j a_j}\right),
    \quad
    a_i' = \begin{cases}
        \lambda w_i a_i, & i\in\mathcal{P},\\
        a_i, & i\notin\mathcal{P},
    \end{cases}
    \quad
    A_i = a_i' - \frac{1}{G}\sum_{j=1}^G a_j'.
    \label{eq:posttrain:coding_agent_gar}
\end{equation}
When the group pass rate is sufficiently high, a bounded length penalty discourages unnecessarily long successful rollouts. Token-level penalties additionally suppress positive credit or amplify negative credit for tool-call or format errors.

\subsection{Model Fusion}
\label{sec:posttrain:fusion}
To efficiently fuse complementary capabilities into a unified model, \modelinstruct{} adopts a two-stage model fusion pipeline that combines weight-space merging with MOPD. Model merging provides a multi-domain initialization for the student, while MOPD further integrates capabilities from specialized RL teachers under the student's own rollout distribution.

\subsubsection{Model Merge}
\label{sec:posttrain:model_merge}

We merge the weights of the non-thinking model and the domain SFT models from \autoref{sec:posttrain:specialist_sft} into a single checkpoint.
This checkpoint initializes the student policy, which then undergoes a lightweight SFT stage on a limited mixed-domain dataset before the first distillation rollout.

\subsubsection{Multi-Teacher On-Policy Distillation}
\label{sec:posttrain:mopd}

Starting from the merged checkpoint, we follow DeepSeek-V4~\citep{deepseekv4} and apply full-vocabulary MOPD to transfer capabilities from more than ten SFT and RL teachers to the student policy.

\textbf{Data Selection:}
Each RL teacher is distilled on inputs drawn from its RL training distribution.
For multi-turn agent data, we follow the prefix replay approach of ReOPD and MiMo MOPD2 and convert recorded trajectories into single-turn examples~\citep{liao2026reopd}. Each example ends immediately before an assistant turn and retains the task prompt and preceding interaction history, including any recorded tool results.

\textbf{Distillation Procedure:}
For each input, the current student policy samples a continuation. The corresponding teacher computes a next-token distribution over the full vocabulary, conditioned on the same input and the tokens already generated by the student. We update the student by minimizing the full-vocabulary reverse KL divergence, $\mathrm{D}_{\mathrm{KL}}(p_{\mathrm{student}}\|p_{\mathrm{teacher}})$, at the newly generated token positions.
Each agent rollout starts from a fixed prefix that reuses recorded tool results and generates one new assistant turn, thereby distillation requires no environment interaction.

% \begin{figure}[!htbp]
%     \centering
%     \includegraphics[width=\textwidth]{figures/posttrain_mopd_agent.png}
%     \caption{Off-environment MOPD for agent tasks. Each recorded trajectory provides a prefix for each assistant turn. The student samples a new turn (orange) from a fixed prefix, and the teacher scores the generated tokens conditioned on the same history.}
%     \label{fig:posttrain:mopd_agent}
% \end{figure}

\subsection{Evaluation}
\label{sec:posttrain:eval}

We comprehensively evaluate \modelname{} against recent open-weight post-trained models across a diverse range of benchmarks, covering reasoning, knowledge, instruction following, writing, coding, and agentic capabilities. 
% Our evaluation suite spans both English and Chinese, as well as multiple programming languages.

\subsubsection{Evaluation Settings}
\label{sec:posttrain:evaluation_settings}

To comprehensively assess the capabilities of \modelname{},
we organize our evaluation suite into two primary categories:
general capabilities and agentic capabilities.

\paragraph{General Capabilities.}
IFBench~\citep{ifbench},
AutoLogi~\citep{autologi},
GPQA Diamond~\citep{gpqa},
AA-Omniscience-v1.1 Public~\citep{aa_omniscience},
HMMT 2026~\citep{hmmt2026},
IMO-AnswerBench-v2~\citep{imo_answerbench},
SciCode~\citep{scicode},
ICPC 2025~\citep{icpc2025},
and IOI 2026~\citep{ioi2026}.

\paragraph{Agentic Capabilities.}
JobBench~\citep{jobbench},
$\tau^3$-Banking~\citep{tau3banking},
$\tau^3$-Bench~\citep{tau3bench},
SWE-Bench Pro~\citep{swebench_pro},
DeepSWE-v1.1~\citep{deepswe},
HLE with tools~\citep{hle},
BrowseComp~\citep{browsecomp},
BrowseComp-ZH~\citep{browsecomp_zh},
WideSearch~\citep{widesearch},
DeepSearchQA~\citep{deepsearchqa},
Agents' Last Exam~\citep{agents_last_exam},
AutomationBench~\citep{automationbench},
SWE-bench Verified~\citep{swebench_verified},
Terminal-Bench 2.1~\citep{terminal_bench},
and BFCL-v4~\citep{bfcl}.

Detailed benchmark specifications and evaluation protocols
are provided in \autoref{app:post_evaluation_details}.

% \paragraph{Evaluation Protocol.}
We conduct evaluations using our internal framework with consistent
protocols across models. For comparison models, we adopt their officially
recommended sampling parameters. For \modelname{}, we set the
temperature to 1.0 and top-$p$ to 0.95.
Unless otherwise specified, we use a context window of 256K tokens.
For agent benchmarks, we adopt benchmark-specific harnesses and
consistent execution budgets across models.
Detailed evaluation protocols, including benchmark configurations,
sampling repetitions, and scoring methods, are provided in
\autoref{app:post_evaluation_details}.

% \paragraph{Mitigating Answer Leakage and Evaluation Exploitation.}
% To improve evaluation reliability, we implement benchmark-specific safeguards against answer leakage and exploitation of the evaluation environment.
% For HLE with tools, we adopt the official CAIS blocklist%
% \footnote{\url{https://github.com/centerforaisafety/hle/blob/main/docs/blocklist.json}},
% excluding eight domain-wide restrictions on general-purpose research platforms such as GitHub and Stack Overflow.
% We retain the remaining 212 domain rules and known leakage-page
% restrictions, and apply filtering to search queries, search results, retrieved webpages, and summaries.
% For DeepSearchQA, we block URLs associated with benchmark datasets, mirrors, repositories, leaderboards, and the benchmark paper, using both benchmark names and dataset-specific aliases.

To improve evaluation reliability, we implement
benchmark-specific safeguards against answer leakage
and exploitation of the evaluation environment.
For HLE with tools, we adopt the official CAIS
blocklist~\citep{hleblocklist}
with minor adaptations to preserve access to
legitimate research resources.
For DeepSearchQA, we apply additional
filtering mechanisms to prevent direct access to
benchmark-specific materials and potential sources
of answer leakage.

% BrowseComp and WideSearch are currently evaluated without equivalent benchmark-specific URL filtering.

% \subsubsection{Evaluation Results}

% \label{sec:evaluation_results}

% Table~\ref{tab:main_results} reports the complete unified evaluation. We mark
% the best result in each row in bold and the second-best result with an
% underline. A dash indicates that a result is unavailable under the specified
% configuration.

% % Because accuracy, Elo, win rate, and rubric-based scores are
% % not commensurable, we do not average all benchmarks into a single overall
% % score. Domain-level macro averages are computed only over compatible
% % percentage-based metrics; Elo and pairwise win rates are analyzed separately.

\begin{table*}[!t]
\centering
\small
\caption[Post-training model comparison]{Performance comparison of \modelinstruct{} against recent open-weight post-trained models across diverse evaluation domains. Scores of \modelinstruct{} are bold when best and underlined when second best. A dash indicates that a result is unavailable.}
\label{tab:main_results}
\setlength{\tabcolsep}{3.5pt}
\renewcommand{\arraystretch}{1.05}

\begin{adjustbox}{max width=\textwidth, max totalheight=0.96\textheight, keepaspectratio}
\begin{tabular}{l|cccc}
\toprule
\textbf{Benchmark}
& \textbf{\modelinstruct{}}
& \begin{tabular}[c]{@{}c@{}}\textbf{DeepSeek-V4}\\\textbf{Flash-0731}\end{tabular}
& \begin{tabular}[c]{@{}c@{}}\textbf{DeepSeek-V4.1}\\\textbf{Flash-1017}\end{tabular}
& \begin{tabular}[c]{@{}c@{}}\textbf{GLM-5.3}\\\textbf{Flash}\end{tabular} \\
\midrule

% -------------------- General Capabilities --------------------
\rowcolor{black!5}
\multicolumn{5}{l}{\rule[-0.5em]{0pt}{1.6em}\large\textbf{General Capabilities}} \\
IFBench & \textbf{73.71} & 73.00 & 73.55 & 59.87 \\
Autologi & \underline{89.51} & 88.81 & 83.89 & 90.68 \\
GPQA Diamond & 90.53 & 89.90 & 92.05 & 90.91 \\
AA-Omniscience-v1.1 Public & \underline{-0.67} & -19.67 & -8.67 & 2.33 \\
HMMT 2026 & \underline{93.18} & 92.61 & 96.59 & 92.42 \\
IMO AnswerBench-v2 & 90.38 & 90.84 & 95.44 & 65.25\rlap{\textsuperscript{\hyperlink{fn:glm_imo}{\ref*{fn:glm_imo}}}} \\
SciCode & \underline{50.93} & 50.30 & 56.60 & 50.23 \\
% AA-LCR-v1.1 & 75.67 & 79.00 & 82.33 & 79.67 \\
ICPC 2025 & \goldmedal{} & -- & -- & -- \\
IOI 2026 & \goldmedal{} & -- & -- & -- \\
% AIME 2026 & 95.83 & 98.54 & 99.58 & 89.17 \\
% ARC-AGI-2 & -- & 61.67 & 80.83 & 41.81 \\
% ZebraLogic & -- & 98.40 & 99.80 & 96.20 \\
% MMLU-Pro & -- & 86.87 & 87.95 & 87.11 \\
% SuperGPQA & -- & 69.15 & 71.95 & 69.38 \\
% IFEval & 91.31 & 94.20 & 95.40 & 88.31 \\
% Multi Challenge & 57.28 & 65.15 & 75.76 & 66.61 \\
% Arena Hard v2 Creative & 76.66 & 91.48 & 94.30 & 84.12 \\
% Creative Writing-v3 & -- & 81.98 & 79.36 & 86.43 \\
% WritingBench & -- & 81.34 & 81.38 & 82.77 \\
% EQ-Bench v3 & -- & 77.95 & 78.75 & 83.85 \\
% LiveCodeBench-v6 & 87.22 & 91.35 & 92.46 & 81.44 \\
% MultiPL-E HumanEval & -- & 88.23 & 90.56 & 89.47 \\
% OJBench C++ & -- & 77.69 & 84.48 & 56.25 \\
% OJBench Python & -- & 68.91 & 77.26 & 44.20 \\

% -------------------- Agentic Capabilities --------------------
\rowcolor{black!5}
\multicolumn{5}{l}{\rule[-0.5em]{0pt}{1.6em}\large\textbf{Agentic Capabilities}} \\
JobBench & \textbf{58.84} & 43.14 & 53.53 & 50.28 \\
$\tau^3$-Banking & \underline{53.61} & 43.30 & 43.30 & 54.64 \\
$\tau^3$-Bench & \underline{81.07} & 80.53 & 78.40 & 84.00 \\
SWE-Bench Pro & \underline{64.30} & 57.64 & 67.40 & 62.20 \\
DeepSWE-v1.1 & \underline{69.03} & 57.52 & 77.00 & 53.09 \\
HLE w/ Tools & 49.23 & 53.65 & 59.56 & 52.90 \\
BrowseComp & \underline{85.50} & 83.00 & 87.50 & 72.00 \\
BrowseComp ZH & \underline{77.80} & 73.40 & 80.97 & 71.62 \\
WideSearch & \underline{79.08} & 78.45 & 79.84 & 73.78 \\
DeepSearchQA & \underline{91.80} & 87.00 & 93.30 & 91.10 \\
Agents' Last Exam & 26.70 & 28.00 & 36.00 & 25.33 \\
AutomationBench & \underline{50.50} & 37.30 & 56.17 & 49.50 \\
SWE-bench Verified & \underline{92.00} & 88.00 & 95.00 & 89.65 \\
Terminal-Bench 2.1 & \underline{85.39} & 83.90 & 86.89 & 75.28 \\
BFCL-v4 & \textbf{71.06} & 70.27 & 69.28 & 70.92 \\
% GDPval (Win Rate) & -- & 21.90 & 51.78 & 49.01 \\
% GDPval (Elo) & -- & 1380.00 & 1613.00 & 1593.00 \\
% GDP.pdf & -- & 13.65 & 16.43 & 10.95 \\
% AA-Briefcase-Lite & -- & 58.18 & 61.80 & 50.90 \\
% Terminal-Bench 4.0 & -- & 6.45 & 10.75 & 32.26 \\

\bottomrule
\end{tabular}
\end{adjustbox}
\end{table*}

\subsubsection{Evaluation Results}

\paragraph{General Capabilities.}\enlargethispage{\baselineskip}
Across general benchmarks, \modelinstruct{} achieves strong performance in reasoning, knowledge, and instruction following.
It leads all compared models on IFBench~(73.71) and ranks second on AutoLogi~(89.51), AA-Omniscience-v1.1~Public~(-0.67), HMMT~2026~(93.18), and SciCode~(50.93), outperforming DeepSeek-V4-Flash-0731 on all of these as well as on GPQA~Diamond~(90.53). It also achieves gold-medal performance at ICPC~2025 and IOI~2026.

\paragraph{Agentic Capabilities.}\stepcounter{footnote}\footnotetext{\label{fn:glm_imo}\raisebox{1.2em}[0pt][0pt]{\hypertarget{fn:glm_imo}{}}The GLM-5.3 Flash API limits the maximum output length to 128K tokens, and 28\% of its responses were truncated.}
\modelinstruct{} is particularly strong on agentic and tool-use benchmarks, outperforming DeepSeek-V4-Flash-0731 on all but two tasks. It leads all compared models on JobBench~(58.84) and BFCL-v4~(71.06), and ranks second on $\tau^3$-Bench~(81.07), $\tau^3$-Banking~(53.61), BrowseComp~(85.50), BrowseComp~ZH~(77.80), WideSearch~(79.08), DeepSearchQA~(91.80), and AutomationBench~(50.50).
In agentic software engineering, it achieves 92.00 on SWE-bench~Verified, 85.39 on Terminal-Bench~2.1, 69.03 on DeepSWE-v1.1, and 64.30 on SWE-Bench~Pro, surpassing DeepSeek-V4-Flash-0731 and GLM-5.3 Flash on all four while remaining competitive with DeepSeek-V4.1-Flash-1017.

\section{Conclusion and Future Directions}
\label{sec:conclusion}

Human cognitive capability is not fully formed from the outset; rather, it gradually constructs more complex structures through continuous learning and adaptation, just as we do not educate young children and university students in the same way.
Inspired by this developmental logic, \modelname{} introduces \emph{PACE}, a training paradigm that flexibly adapts the learning pace to the model's evolving cognitive capacity.
At its core, PACE coordinates three synergistic pillars: \emph{InfiGrow} expands model width and depth in stages as learning slows, \emph{InfiCoScale} shifts the training distribution toward richer, harder, and higher-quality data in step with the newly added capacity, and \emph{InfiEPFlow} provides communication-aware expert-parallel load balancing that significantly reduces communication costs.
Through this synergy, \modelname{} achieves strong frontier-level performance at a fraction of conventional training cost, delivering roughly $8\times$ the \emph{intelligence per compute}~(IPC) of the next most efficient compared model and demonstrating that high-capability foundation models need not be trained at full scale from day one.
Specifically, \modelbase{} exceeds DeepSeek-V4-Flash-Base and MiMo-V2.5-Base in pre-training, and after post-training, \modelinstruct{} surpasses DeepSeek-V4-Flash~(0731) while rivaling DeepSeek-V4.1-Flash~(1017) across a range of domains.

Looking ahead, we will continue advancing the PACE paradigm toward fine-grained scheduling of model growth and data co-scaling, tighter integration with low-precision training, and more efficient training infrastructure to further lower the compute and hardware barriers for frontier intelligence.

\bibliography{main}

\newpage
\appendix

\section*{Appendix}

\section{Author List and Acknowledgment}
\label{app:authors}

We sincerely thank all contributors for their invaluable support and dedication to this work.
Authors within each role are listed alphabetically by their first name.

\vspace{1em}
\noindent\textbf{Leads}

\vspace{0.5em}
{\color[HTML]{1F2D7A}
\noindent\begin{tabular}{@{}p{0.48\textwidth}p{0.48\textwidth}@{}}
Jianmin Wu & Hongxia Yang \\
\end{tabular}}

\vspace{1em}
\noindent\textbf{Contributors}

\vspace{0.5em}
{\color[HTML]{1F2D7A}
\noindent\begin{tabular}{@{}p{0.48\textwidth}p{0.48\textwidth}@{}}
Botao Yang & Weiqian Lin \\
Cheng Li & Wenjun Wang \\
Chenghao Zhang & Xijun Xiang \\
Congkai Xie & Xinxin Peng \\
Dongkuan Wu & Xuan Ma \\
Donglai Ge & Xuhuan Li \\
Fangwei Zuo & Ya Gao \\
Gengqi Yang & Yanggan Gu \\
Hanjie Fu & Yifan Yang \\
Hao Ruan & Yijin Zhou \\
Hao Xue & Yiming Zhang \\
Jihang Zhang & Youhui Bai \\
Jingwei Shi & Yuanyi Wang \\
Jinjia Li & Yueyang Xing \\
Kejing Yang & Yuhang Liu \\
Lei Chen & Yujie He \\
Ming Li & Yujuan Cao \\
Ouxiang Zhou & Yuqin Huang \\
Pan Sun & Yuyang Cheng \\
Pengxiang Li & Yuzhe Li \\
Qiang Xu & Zhan Chen \\
Ran Liu & Zhaoyi Yan \\
Shizhan Fang & Zhen Hu \\
Shuyue Deng & Zheng Liu \\
Songmiao Wang & Zhenquan Yao \\
Su Lu & Zhichao Zheng \\
Tian Wang & Zhijie Sang \\
Tianyi Lyu & Zhitian Hou \\
Tong Bai & \\
\end{tabular}}

% \section{Evaluation Details}
% \label{app:eval}

\section{Growth Scaling Law}
\label{app:upscale}

\paragraph{Model configurations.}
All models in the growth scaling law study follow the same design rules, so that the comparison across scales is not confounded by architectural choices.
The MoE configuration is fixed across scales: every model has 64 routed experts and 1 shared expert, with 6 routed experts activated per token.
The first layer uses a dense FFN with an intermediate size of $\frac{8}{3}d_{\mathrm{model}}$, and all remaining layers are MoE layers.
All attention layers use full attention with GQA and a head dimension of 128.
The remaining dimensions are chosen within the ranges where the loss is insensitive to the model shape according to scaling laws: the ratio $d_{\mathrm{model}}/n_{\mathrm{layers}}$ of the original models lies between 60 and 100, and in each MoE layer, the activated routed-expert parameters are 2 to 3 times the attention parameters~\citep{kaplan2020scaling}.
Since the grown models keep the width of the original models, their $d_{\mathrm{model}}/n_{\mathrm{layers}}$ is lower.
The resulting configurations are listed in \autoref{tab:upscale_models}.

\paragraph{Training details.}
All runs use the same data, a sequence length of 8K and a global batch size of 512 sequences, i.e., about 4.2M tokens per step.
All models are optimized with AdamW~\citep{adamw} with $\beta_1 = 0.9$, $\beta_2 = 0.95$ and a weight decay of 0.1, and all parameters are initialized with a standard deviation of 0.006.
The learning rate is linearly warmed up to $10^{-3}$ over the first 1,000 steps and then kept constant.
Decay runs are branched from the constant-learning-rate checkpoints at about 17B, 34B, 67B and 101B tokens, and each decays the learning rate from $10^{-3}$ to $10^{-4}$ with a cosine schedule over 10\% of the total training steps, which gives one annealed loss per token budget at the cost of a single run plus about 10\% extra steps for each budget.
After each depth growth, the learning rate is warmed up again over 100 steps.

\begin{table}[!t]
    \centering
    \caption{
        Model configurations used in the growth scaling law study.
        Each pair consists of an original model and its depth-grown counterpart, which keeps the width of the original model and only increases the number of layers.
    }
    \label{tab:upscale_models}
    \setlength{\tabcolsep}{3.5pt}
    \begin{tabular*}{\textwidth}{@{\extracolsep{\fill}}l l c c c c@{}}
    \toprule
    & \textbf{Model} & \textbf{Layers} & \textbf{Hidden} & \textbf{Heads (Q/KV)} & \textbf{Expert FFN} \\
    \midrule
    \multirow{2}{*}{Pair 1} & 0.9B-A130M & 12 & 1024 & 8/2 & 384 \\
    & 1.4B-A200M & 18 & 1024 & 8/2 & 384 \\
    \midrule
    \multirow{2}{*}{Pair 2} & 2B-A280M & 16 & 1280 & 10/2 & 512 \\
    & 3B-A430M & 24 & 1280 & 10/2 & 512 \\
    \bottomrule
    \end{tabular*}
\end{table}

\section{Hybrid Attention Ablations}
\label{app:ablation}

The ablation studies use a 16B-A3B MoE model, whose configuration is listed in \autoref{tab:ablation_model_config}.
The first layer and the last two layers use FA, and the remaining 24 layers repeat one FA layer followed by three SWA layers.
All attention layers use QK-norm and a learnable attention sink, and RoPE is applied to the first 64 dimensions of query and key.

\begin{table}[!t]
    \centering
    \small
    \caption{Configuration of the 16B-A3B MoE model used in the ablation studies.}
    \label{tab:ablation_model_config}
    \begin{tabular}{@{}l l c@{}}
    \toprule
    \textbf{Block} & \textbf{Configuration} & \textbf{Value} \\
    \midrule
    \multirow{2}{*}{\textbf{Backbone}}
    & Layers (Total/SWA/FA) & 27/18/9 \\
    & Hidden Size & 2048 \\
    \midrule
    \multirow{5}{*}{\textbf{Attention}}
    & SWA Heads (Q/KV) & 16/8 \\
    & SWA Head Dimensions (QK/V) & 192/128 \\
    & Sliding Window Size & 512 \\
    & FA Heads (Q/KV) & 16/4 \\
    & FA Head Dimensions (QK/V) & 128/128 \\
    \midrule
    \multirow{5}{*}{\textbf{MoE}}
    & Layers (Dense/MoE) & 1/26 \\
    & Experts (Routed/Shared/Activated) & 64/1/6 \\
    & Expert FFN Dimension & 1408 \\
    & Shared Expert FFN Dimension & 2816 \\
    & Dense FFN Dimension & 11,264 \\
    \bottomrule
    \end{tabular}
\end{table}

\paragraph{Hybrid attention.}
\autoref{tab:ablation_swa} shows a series of experiments that differ only in their attention configurations.
Increasing the number of KV heads in SWA layers (SWA-1) brings the hybrid model close to the full-attention baseline FA, and further enlarging the head dimension (SWA-2) even surpasses it.
Moreover, SWA-3, which enlarges the head dimension only in SWA layers, performs comparably to SWA-2, indicating that a larger head dimension in FA layers brings little additional gain.
Increasing the number of query heads in SWA layers together with gated attention (SWA-4) shows no obvious gain over simply increasing the KV heads in SWA layers (SWA-1).

\begin{table}[!t]
    \centering
    \caption{
        FA vs.\ SWA. Both models are evaluated after 80B training tokens.
        FA: every layer is an FA layer. SWA: FA and SWA layers interleaved at a ratio of 1:3 with the same head configuration, with an attention sink in every layer.
    }
    \label{tab:ablation_fa_swa}
    \setlength{\tabcolsep}{3.5pt}
    \begin{tabular*}{\textwidth}{@{\extracolsep{\fill}}l c c c c c c c c@{}}
    \toprule
    \textbf{Model} & MMLU & MMLU-Pro & BBH & CMMLU & GSM8K & HumanEval & MBPP & Avg. \\
    \midrule
    FA & 45.8 & 19.4 & 33.4 & 32.3 & 30.2 & 15.8 & 35.0 & 30.3 \\
    SWA & 42.1 & 17.6 & 32.1 & 30.9 & 25.9 & 17.7 & 37.4 & 29.1 \\
    \bottomrule
    \end{tabular*}
\end{table}

\begin{table}[!t]
    \centering
    \caption{
        Ablation on the hybrid attention design.
        All models are trained on the same 90B tokens with a sequence length of 8K and a global batch size of 512, and differ only in their attention configurations.
        Heads are given as query heads/KV heads.
        FA: full-attention baseline, in which every layer is an FA layer with an attention sink.
        SWA-1 to SWA-4 interleave FA and SWA layers at a ratio of 1:3.
        SWA-1: 8 KV heads in SWA layers.
        SWA-2: SWA-1 with a head dimension of 192 in all layers.
        SWA-3: SWA-1 with a head dimension of 192 in SWA layers and 128 in FA layers.
        SWA-4: 24 query heads in SWA layers, using head-wise gated attention~\citep{gatedattn} instead of the attention sink.
    }
    \label{tab:ablation_swa}
    \setlength{\tabcolsep}{3.5pt}
    \begin{tabular*}{\textwidth}{@{\extracolsep{\fill}}l l | c c c c c@{}}
    \toprule
    & \textbf{Benchmark} & \textbf{FA} & \textbf{SWA-1} & \textbf{SWA-2} & \textbf{SWA-3} & \textbf{SWA-4} \\
    \midrule
    \multirow{4}{*}{Attention} & FA Heads & 16/4 & 16/4 & 16/4 & 16/4 & 16/4 \\
    & FA Head Dimension & 128 & 128 & 192 & 128 & 128 \\
    & SWA Heads & - & 16/8 & 16/8 & 16/8 & 24/4 \\
    & SWA Head Dimension & - & 128 & 192 & 192 & 128 \\
    \midrule
    \multirow{4}{*}{English} & MMLU & 47.1 & 45.1 & 46.7 & 47.8 & 44.2 \\
    & MMLU-Pro & 19.1 & 17.6 & 20.7 & 19.8 & 16.6 \\
    & BBH & 28.3 & 29.7 & 29.9 & 29.1 & 29.2 \\
    & TriviaQA & 53.7 & 55.1 & 55.4 & 55.1 & 54.2 \\
    \midrule
    \multirow{3}{*}{Chinese} & CMMLU & 49.0 & 47.8 & 49.0 & 50.1 & 46.4 \\
    & C-Eval & 46.7 & 47.9 & 48.4 & 47.5 & 49.7 \\
    & C3 & 70.2 & 68.5 & 71.5 & 73.0 & 65.1 \\
    \midrule
    \multirow{2}{*}{Math} & GSM8K & 22.6 & 18.4 & 24.3 & 22.4 & 20.6 \\
    & MATH & 10.7 & 10.1 & 6.9 & 10.4 & 11.6 \\
    \midrule
    \multirow{2}{*}{Code} & HumanEval & 17.1 & 12.8 & 15.2 & 15.8 & 13.4 \\
    & MBPP & 27.2 & 26.5 & 29.6 & 30.7 & 31.5 \\
    \midrule
    & Average & 35.6 & 34.5 & 36.2 & 36.5 & 34.8 \\
    \bottomrule
    \end{tabular*}
\end{table}

\begin{table}[!t]
    \centering
    \caption{
        Ablation on width and depth growth, evaluated after 160B training tokens.
        Baseline: the model before growth, trained without growth.
        Depth: depth growth by stacking.
        Width-1: increasing the number of activated experts by 2.
        Width-2: duplicating the 8 hottest experts and increasing the number of activated experts by 2.
        Width-3: duplicating all routed experts and increasing the number of activated experts by 6.
    }
    \label{tab:ablation_upscale}
    \setlength{\tabcolsep}{3.5pt}
    \begin{tabular*}{\textwidth}{@{\extracolsep{\fill}}l l | c c c c c@{}}
    \toprule
    & \textbf{Benchmark} & \textbf{Baseline} & \textbf{Width-1} & \textbf{Width-2} & \textbf{Width-3} & \textbf{Depth} \\
    \midrule
    \multirow{3}{*}{Model} & Layers & 27 & 27 & 27 & 27 & 43 \\
    & Routed Experts & 64 & 64 & 72 & 128 & 64 \\
    & Activated Experts & 6 & 8 & 8 & 12 & 6 \\
    \midrule
    \multirow{4}{*}{English} & MMLU & 51.2 & 51.6 & 49.5 & 52.3 & 57.4 \\
    & MMLU-Pro & 24.5 & 23.6 & 23.0 & 22.7 & 28.3 \\
    & BBH & 35.6 & 38.5 & 38.3 & 37.5 & 43.2 \\
    & TriviaQA & 59.6 & 59.2 & 59.5 & 59.5 & 59.7 \\
    \midrule
    \multirow{3}{*}{Chinese} & CMMLU & 55.8 & 55.0 & 55.7 & 56.4 & 60.2 \\
    & C-Eval & 54.7 & 54.8 & 54.3 & 55.7 & 58.2 \\
    & C3 & 74.8 & 77.4 & 76.1 & 77.7 & 84.1 \\
    \midrule
    \multirow{2}{*}{Math} & GSM8K & 36.2 & 36.9 & 39.3 & 37.1 & 50.1 \\
    & MATH & 15.5 & 18.8 & 17.7 & 20.4 & 22.7 \\
    \midrule
    \multirow{2}{*}{Code} & HumanEval & 17.7 & 18.9 & 20.7 & 18.9 & 22.6 \\
    & MBPP & 31.5 & 33.1 & 36.2 & 33.9 & 41.2 \\
    \midrule
    & Average & 41.5 & 42.5 & 42.8 & 42.9 & 48.0 \\
    \bottomrule
    \end{tabular*}
\end{table}

% \input{tables/ablation_width_depth}

% prompt n-gram去重，训练加入固定pattern，可能有明显的下降
% token对齐
\section{Pretrain Evaluation Settings \& Prompt Examples}
\label{app:benchmark_prompt}

Benchmark scores are sensitive to implementation details, especially for base models: with weaker instruction-following, they rely heavily on demonstrations to establish the output format and on the framework to extract the final answer for evaluation. We therefore carefully tune the evaluation framework, select the number of demonstrations and the response budget for each benchmark, and use the same settings for all models.

\subsection{Prompt Design}

\paragraph{CoT.}
Each benchmark forces a choice of prompt style, and neither choice is free. A reasoning trace gives a base model room to work through a problem, but on tasks that only require recall or reading it invites detours and makes the final answer harder to extract reliably. A short-answer prompt keeps the answer clean, but on a problem that takes several steps of derivation it invites an early guess. We therefore apply chain-of-thought where the answer requires several steps of derivation, and keep short-answer prompts for tasks that require recall or reading.

\paragraph{Few-shot examples.}
The number of demonstrations is fixed per benchmark across all models, including MMLU-Pro~\citep{mmlupro}, DROP~\citep{drop}, BBH~\citep{bbh}, GPQA-Diamond~\citep{gpqa} and the math benchmarks~\citep{gsm8k,math}. Because a base model follows demonstrations rather than explicit output instructions, its format largely determines whether the final answer can be extracted at all. The prompts, therefore, share a template but not a reasoning style; the examples below illustrate the difference.

% \paragraph{Prompt boundary.}
% A base model continues from the last token, not from the last character, so the exact
% ending of a template sets the boundary from which generation starts. In most BPE
% tokenizers, a space before a word belongs to that word's token, so a query ending in a
% bare space leaves that space as a standalone token instead of matching the
% demonstrations' boundary. More generally, ending on a prefix of a longer token forces a
% split rarely seen in training, and the model assigns little probability to the
% continuation that would complete the merged
% token~\citep{hayase2025pbp,xu2026partialtoken}. We therefore end every query at its
% response marker (e.g., the \texttt{Answer:} line) just as the demonstrations do, verify
% the boundary of each rendered prompt with the tokenizer of the model under evaluation,
% and fix any mismatches in the template. The rendered string is held fixed across
% checkpoints so that score differences reflect the model rather than the prompt.
\paragraph{Prompt boundary.}
Where a prompt ends affects a base model's score, including differences as small as a trailing space.
The cause lies in tokenization: in most BPE tokenizers a space before a word belongs to that word's token, so \texttt{Answer:} and \texttt{Answer:\textvisiblespace} leave the model in two different states.
After the former it continues with a normal token such as \texttt{\textvisiblespace 42}; after the latter the trailing space has become a token of its own, and the model must continue with \texttt{42} alone.
This is a split it has rarely seen in training, and it assigns little probability to the continuation that would complete the merged token~\citep{hayase2025pbp,xu2026partialtoken}.
The same holds whenever a prompt ends on a prefix of a longer token.
A few-shot prompt imposes a further constraint: the demonstrations show the model where an answer begins, and a query that ends differently breaks the very pattern it is meant to follow.
We therefore end every query exactly as the demonstrations do, at the response marker itself, verify the boundary of each rendered prompt with the tokenizer of the model under evaluation, and fix any mismatch in the template.
The rendered string is then held fixed across checkpoints, so that score differences reflect the model rather than the prompt.

\subsection{Prompt Examples}
The prompts below cover multiple-choice knowledge, reading comprehension, code reasoning, and mathematics.
Each shows a complete prompt built around a real benchmark item; the reasoning traces are worked solutions written to the benchmark's answer and are shown in full.

\paragraph{MMLU-Pro.}
MMLU-Pro~\citep{mmlupro} uses 5-shot prompting with shots drawn from the validation split; the listing shows one shot, and the query is a placeholder test item. The query ends at \texttt{Answer: Let's think step by step.}, generation is capped at 1,024 tokens, and generation stops at \texttt{Question:\textbackslash n}.
\begin{lstlisting}[basicstyle=\ttfamily\scriptsize, frame=single, breaklines=true, columns=fullflexible]
The following are multiple choice questions (with answers) about biology. Think
step by step and then finish your answer with "the answer is (X)" where X is the
correct letter choice.
Question:
Which of the following represents an accurate statement concerning arthropods?
Options:
A. They possess an exoskeleton composed primarily of peptidoglycan.
B. They possess an open circulatory system with a dorsal heart.
C. They are members of a biologically unsuccessful phylum incapable of
exploiting diverse habitats and nutrition sources.
D. They lack paired, jointed appendages.
Answer: Let's think step by step. Peptidoglycan is known to comprise the plasma
membrane of most bacteria, rather than the exoskeleton of arthropods, which is
made of chitin, which rules out (A). The answer (C) is false because arthropods
are a highly successful phylum. Likewise, arthropods have paired, jointed
appendages, which rules out (D). The only remaining option is (B), as arthropods
have an open circulatory system with a dorsal tubular heart. The answer is (B).

<Four further shots...>

Question:
<test question>
Options:
A. <option>
B. <option>
C. <option>
D. <option>
E. <option>
F. <option>
G. <option>
H. <option>
I. <option>
J. <option>
Answer: Let's think step by step.
\end{lstlisting}

\paragraph{DROP.}
DROP~\citep{drop} uses 5-shot prompting; the listing shows one shot; the rendered query ends with the newline after \texttt{\#\#\# Reasoning:}, generation is capped at 1,024 tokens, and generation stops at \texttt{\textbackslash n\#\#\# Passage:} or \texttt{\textbackslash n\#\#\# Question:}.
\begin{lstlisting}[basicstyle=\ttfamily\scriptsize, frame=single, breaklines=true, columns=fullflexible]
### Passage: The Croatian Home Guard reached its maximum size at the end of 1943,
when it had 130,000 men. It also included an air force, the Air Force of the
Independent State of Croatia, the backbone of which was provided by 500 former
Royal Yugoslav Air Force officers and 1,600 NCOs with 125 aircraft. By 1943 the
ZNDH was 9,775 strong and equipped with 295 aircraft.
### Question: Where did most of the Air Force come from?
Let's think step by step.
### Reasoning:
It came from the former Royal Yugoslav Air Force; the passage says these officers
provided the backbone.

## Summary:
former Royal Yugoslav Air Force

<Four further shots...>

### Passage: <test passage>
### Question: <test question>
Let's think step by step.
### Reasoning:
\end{lstlisting}

\paragraph{CRUXEval.}
CRUXEval~\citep{cruxeval} averages two subsets: CRUXEval-I (predict an input from an output) and CRUXEval-O (predict an output from an input). Both use 2-shot prompting with a format instruction and two fixed demonstrations in a \texttt{\#\#\# Question / \#\#\# Reasoning / \#\#\# Answer} format. The model reasons in the \texttt{\#\#\# Reasoning} section and returns a passing assertion in the \texttt{\#\#\# Answer} section. The rendered prompt ends with the newline after \texttt{\#\#\# Reasoning}; generation is capped at 2,048 tokens; the configured stop strings, \texttt{\textbackslash n\#\#\# Question} and \texttt{\textbackslash n\#\#\# Reasoning}, are the two section headings and apply only to the generated continuation; the returned assertion is executed against the expected output (Pass@1).

\textbf{CRUXEval-I:}
The model receives a function and an output and predicts an input; the listing shows one of the two shots and a placeholder test case.
\begin{lstlisting}[basicstyle=\ttfamily\scriptsize, frame=single, breaklines=true, columns=fullflexible]
You will be given a function f and an output in the form f(??) == output. Your
task is to find any input such that executing f on the input leads to the
given output. There may be multiple answers, but only output one. Work
backwards from the output to construct an input. First, think step by step.
Put the final answer in the ### Answer section as a passing assertion
containing the input and the given output.

### Question
def f(my_list):
    count = 0
    for i in my_list:
        if len(i) % 2 == 0:
            count += 1
    return count
assert f(??) == 3

### Reasoning
To find an input such that executing f on the input leads to 3, we can work
backwards. The function counts elements whose length is even, so we need
exactly three even-length strings. The list ["mq", "px", "zy"] works because
all three strings have length 2.

### Answer
assert f(["mq", "px", "zy"]) == 3

<1 further shot...>

### Question
<test function>
assert f(??) == <test output>

### Reasoning
\end{lstlisting}

\textbf{CRUXEval-O:}
The model receives a function and an input and predicts the output; the listing shows one of the two shots and a placeholder test case.
\begin{lstlisting}[basicstyle=\ttfamily\scriptsize, frame=single, breaklines=true, columns=fullflexible]
You are given a Python function and an assertion containing an input to the
function. Complete the assertion with a literal (no unsimplified expressions,
no function calls) containing the output when executing the provided code on
the given input, even if the function is incorrect or incomplete. Do NOT
output any extra information. Execute the program step by step before
arriving at an answer. Put the full assertion with the correct output in the
### Answer section.

### Question
def f(n):
    return n
assert f(17) == ??

### Reasoning
n is 17, so the function returns n directly, which is 17.

### Answer
assert f(17) == 17

<1 further shot...>

### Question
<test function>
assert f(<test input>) == ??

### Reasoning
\end{lstlisting}

\paragraph{MATH.}
MATH~\citep{math} uses 4-shot prompting with examples drawn from algebra, geometry, and series; the query ends at \texttt{Answer:}, generation is capped at 512 tokens, and generation stops at \texttt{Question:}. The rendered prompt ends at the colon; the shots continue with a space and ``Let's think step by step''.
\begin{lstlisting}[basicstyle=\ttfamily\scriptsize, frame=single, breaklines=true, columns=fullflexible]
Question: What is the area of the region defined by the equation
$x^2+y^2 - 7 = 4y-14x+3$?
Answer: Let's think step by step
We rewrite the equation as $x^2 + 14x + y^2 - 4y = 10$ and then complete the
square, resulting in $(x+7)^2-49 + (y-2)^2-4=10$, or $(x+7)^2+(y-2)^2=63$.
This is the equation of a circle with center $(-7, 2)$ and radius $\sqrt{63}$,
so the area of this region is $\pi r^2 = \boxed{63\pi}$.
The answer is 63\pi

<Three further shots...>

Question: <test problem>
Answer:
\end{lstlisting}

The template is fixed; the reasoning is not.
We intentionally vary the demonstrations' solution paths because identical traces cause the base model to repeat the template, while the answer anchor keeps every output parseable.

% Required packages: booktabs, longtable, array, xcolor (table option), ragged2e
% For a two-column manuscript, switch to \onecolumn before this appendix.
% TODO markers are intentional: confirm them before publication.

\section{Post-training Evaluation Details}
\label{app:post_evaluation_details}

This appendix details the evaluation protocols used in our internal evaluation framework. For benchmarks evaluated over multiple independent runs, we report the mean score. Benchmark-specific configurations complement the general evaluation settings described in \autoref{sec:posttrain:evaluation_settings}.

% ============================================================
% General Capabilities
% ============================================================

\subsection{General Capabilities}

% \paragraph{General Benchmark Protocols.}
We evaluate IFBench over 8 independent runs using rule-based verification
and report strict accuracy. AutoLogi is evaluated with rule-based scoring.
GPQA Diamond uses answer extraction and accuracy-based scoring over 8
independent runs. For AA-Omniscience-v1.1 Public, we evaluate the public
subset of 300 problems and report the net score, using GPT-5.6-Luna at medium
reasoning effort as judge. HMMT 2026 is evaluated over 8 independent runs on
the full problem set using rule-based answer verification. IMO AnswerBench-v2
is evaluated over 4 independent runs with DeepSeek-V4.1-Flash-1017 as judge.
SciCode evaluates 65 problems comprising 288 subproblems over 3 independent
runs, and we report subproblem-level execution accuracy.

\paragraph{ICPC 2025 and IOI 2026.}
We evaluate all 12 problems from the 2025 ICPC World
Finals and all 6 problems from IOI 2026.
Inspired by GenCorrect~\citep{ficek2026posttraining},
we adopt an iterative generation and refinement
protocol with up to five rounds per problem.
In each round, we generate 200 candidate solutions
and select up to 10 submissions based on compilation
success, problem-provided sample test performance, and solution
diversity. Subsequent rounds incorporate selected
solutions and judging feedback from previous rounds.
We cap the total number of submissions at 50 per
problem for both competitions, in accordance with the
official IOI submission limit.

\subsection{Agentic Capabilities}
\label{app:post_agent_configs}

\paragraph{Tool Use and Interactive Tasks.}
For $\tau^3$-Banking and $\tau^3$-Bench, we evaluate airline (50), retail
(114), telecom (114), and banking-knowledge (97) tasks over 4 trials. Each
trial allows up to 200 agent steps and 16K output tokens per generation.
DeepSeek-V4.1-Flash-1017 serves as both user simulator and natural-language
judge. Banking retrieval uses all tools with \texttt{text-embedding-3-large}.
BFCL-v4 evaluates single- and multi-turn function calling, function relevance,
web search, and memory management under the official weighted scoring
protocol, using Serper for search and Jina Reader for retrieval.

\paragraph{Search and Research Agents.}
For HLE with Tools, we evaluate all 2,158 text-only problems with up to 100
agent steps. The reader accepts up to 95,000 input characters and generates
up to 32K tokens. DeepSeek-V4.1-Flash-1017 serves as the judge with medium
reasoning effort, a temperature of 0, and a maximum output length of 8K
tokens. For BrowseComp, we randomly sample 200 problems, while retaining
all 289 BrowseComp-ZH problems, with up to 200 agent steps per problem.
For WideSearch, we evaluate all 200 problems with up to 50 agent steps,
prompting the agent to finalize its answer as the context length
approaches 220K tokens. For DeepSearchQA, we randomly sample 100 problems
with up to 100 agent steps per problem. For BrowseComp, BrowseComp-ZH, and DeepSearchQA,
we adopt a discard-all context management strategy
that retains only the system prompt and the original
question when the estimated input length reaches
80\% of the 256K-token context window.
BrowseComp, BrowseComp-ZH, WideSearch, and DeepSearchQA use Serper and
Jina Reader for retrieval, and DeepSeek-V4.1-Flash-1017 for webpage
summarization and judging at a temperature of 0.

\paragraph{General Agents.}
For JobBench, we use a 256K-token context window with context compression
triggered at 192K tokens.
DeepSeek-V4.1-Flash-1017 serves as the judge in place of Grok-4.3.
Agents' Last Exam evaluates the 75-task text-only ALE v1 subset using OpenClaw,
a maximum of 64K output tokens per generation, and official task scorers with
DeepSeek-V4.1-Flash-1017 for LLM judging when needed. AutomationBench evaluates
6 public domains in the official API simulation environment with up to 50 agent
steps; success requires all assertions to pass.

\paragraph{Coding Agents.}
SWE-Bench Pro uses the OpenHands harness with up to 500 iterations and 32K
output tokens per generation. DeepSWE-v1.1 uses mini-SWE-agent with 48K output
tokens per generation. SWE-bench Verified evaluates all 500 tasks using OpenHands
with up to 250 iterations and 32K output tokens per generation. These three
benchmarks report the resolved rate. Terminal-Bench 2.1 evaluates all 89 tasks
with Codex CLI, a 2-hour budget, up to 500 generation requests, and 32K output
tokens per request. Context compaction starts near 192K tokens. Image viewing,
web search, and multi-agent execution are disabled. Each sandbox receives
4 CPU cores, 32 GiB memory, and 30 GiB storage.

\end{document}